\documentclass[conf]{new-aiaa}
\usepackage[utf8]{inputenc}

\usepackage{graphicx}
\usepackage{amsmath}
\usepackage[version=4]{mhchem}
\usepackage{siunitx}
\usepackage{longtable,tabularx}

\usepackage{relsize}
\usepackage{subfig}

\usepackage{bm}
\usepackage{dsfont}
\usepackage{amssymb}

\usepackage{mathtools}

\usepackage[utf8]{inputenc}
\usepackage[english]{babel}

\usepackage{algpseudocode}
\usepackage{algorithm}
\usepackage[capitalize]{cleveref}
\usepackage{graphicx}
\usepackage{amsmath}

\usepackage{amsthm}
\usepackage[version=4]{mhchem}
\usepackage{siunitx}
\usepackage{longtable,tabularx}
\usepackage{booktabs}

\newtheoremstyle{mystyle}
  {}
  {}
  {\itshape}
  {}
  {\bfseries}
  {.}
  { }
  {\thmname{#1}\thmnumber{ #2}\thmnote{ (#3)}}

\theoremstyle{mystyle}

\newtheorem{remark}{Remark}

\usepackage{subfig}

\renewcommand{\epsilon}{{\varepsilon}}

\usepackage{centernot}
\usepackage{mathtools}
\usepackage{stmaryrd}

\makeatletter
\newcommand{\xMapsto}[2][]{\ext@arrow 0599{\Mapstofill@}{#1}{#2}}
\def\Mapstofill@{\arrowfill@{\Mapstochar\Relbar}\Relbar\Rightarrow}
\makeatother

\usepackage{cancel}
\usepackage{lscape}
\usepackage{nccmath}
\usepackage{rotating}
\usepackage{lineno}

\title{A Tensor-based Structural Health Monitoring Approach for Aeroservoelastic Systems}

\author{Prasad Cheema\footnote{Postgraduate Student, School of AMME}}\affil{The University of Sydney, NSW 2006, Australia}

\author{Nguyen Lu Dang Khoa\footnote{Senior Research Scientist, CSIRO}} \affil{Data61, NSW, CSIRO, 2015, Australia}

\author{Moray Kidd \footnote{Senior Lecturer, School of MAC}} \affil{The University of Manchester, M13 9PL, United Kingdom}

\author{Gareth A. Vio \footnote{Senior Lecturer, School of AMME, SMAIAA}} \affil{The University of Sydney, NSW 2006, Australia}

\begin{document}

\maketitle

\begin{abstract}
Structural health monitoring is a condition-based field of study utilised to monitor infrastructure, via sensing systems. It is therefore used in the field of aerospace engineering to assist in monitoring the health of aerospace structures. A difficulty however is that in structural health monitoring the data input is usually from sensor arrays, which results in data which are highly redundant and correlated, an area in which traditional two-way matrix approaches have had difficulty in deconstructing and interpreting. Newer methods involving tensor analysis  allow us to analyse this multi-way structural data in a coherent manner. In our approach, we demonstrate the usefulness of tensor-based learning coupled with for damage detection, on a novel $N$-DoF Lagrangian aeroservoelastic model. 
\end{abstract}

\section{Introduction}
\lettrine{S}{tructural} health monitoring (SHM) is a complex field of study making use of statistics and robust optimisation in order to allow for damage detection, localization and estimation, through the use of sensing systems. The potential for life-safety and economic benefits has motivated the needs for SHM research, facilitating a shift from time-based to condition-based maintenance \cite{farrar2007introduction}.

Traditionally, vibration-based techniques have been used extensively in SHM for damage identification \cite{alamdari2015}. The time-based response of a structure can be measured by sensors such as accelerometers or strain gauges. Traditional SHM approaches adopt a numerical model, and a physical model of the structure and attempts to relate any differences between the measured data and the data generated by the model as damage identification \cite{worden2007application}. However, a numerical model is not always available in practice and does not always correctly capture the exact behaviour of the real structure. By using statistical and data-driven approaches it is possible to learn a model with confidence bounds from measured data, leading to a more flexible approach to SHM damage identification \cite{farrar2007introduction,worden2007application}.

The key underlying problem in SHM is identifying damage, and it is classified by Rytter into four different levels of complexity \cite{rytter1993}: damage detection , damage localization, damage severity assessment and failure prediction (remaining life estimation). Typically, level 4 necessitates the knowledge of domain-based knowledge of the characteristics of the system and its damage progression. Machine learning methods are typically sufficient to address levels from 1 to 3, with level 1 being traditionally solvable by using an unsupervised learning scheme, while levels 2 and 3 usually necessitate a supervised learning approach \cite{worden2007application,worden2000damage,chan1999}. However recently there have been attempts to approach sub-problem 1 in a supervised manner through the generation of artificial negative data, in order to give more robust bounds for previous unsupervised learning approaches. \cite{Cheema2016_SHM}

Thus clearly, SHM is a field which benefits many industries, and as a result a recent trend in the aviation industry has seen a significant push for the implementation of machine learning techniques in manufacturing, operations and customer satisfaction.These techniques present the potential for predictive maintenance, prognostic component monitoring and aircraft health monitoring, all of which could dramatically reduce the cost of delays and unexpected events on the aviation industry. Accompanying this financial motive, these techniques present the opportunity for an industry wide increase in aircraft operational safety. \cite{lindgren2018us} 

Many major stakeholders in commercial aircraft manufacturing now offer a data based health monitoring system as part of the support package for many components. Airframe manufacturers have extended the scope of their round the clock Aircraft on Ground (AOG) support desks to include the function of a health monitoring system. On their new generation aircraft (A350 \& A380) Airbus offer Aircraft Real Time Health Monitoring as part of their support service. This involves anomaly detection algorithms, classification algorithms and prognostic trend prediction algorithms to monitor the health of an aircraft before departure, in flight and post arrival \cite{AIRTHM}. This service is being developed into a new product marketed as \textit{Skywise} in which the algorithms are expanded to operate more autonomously and over a larger scope of the aircraft's operation. A similar all encompassing service is offered by Boeing on their new generation aircraft \cite{BOEINGh}. Supplemental services are also now offered by the larger suppliers specific to their components. Prognostic algorithms are particularly focused on by the engine manufacturers which aim to predict the remaining useful life of components which operate in the extreme environment of the power plant. These services are being developed across all of the major engine manufacturers including Rolls Royce, Pratt and Whitney and General Electric \cite{ROLLSROYCE, Pratt, GE}. It is important to note that a majority of the functions of these systems are in the process of maturing and only once a critical amount of operational data is provided to them will they become robust. The demand for training data is being met with attempts by the manufacturers to enter non disclosure agreements with the airlines allowing companies like Airbus and Boeing access to airline operational data. However some resistance to this has been experienced from the airlines who have concerns regarding privacy and loss of competitive edge \cite{IATA}.

Although the methods required for addressing SHM problems are well understood, the practical considerations behind SHM are often neglected. In particular it is often required to pool information from multiple sensors in a robust manner, in order to maximise the ability for detecting damage. One such method for sensor fusion may be performed via tensor analysis, which has been successfully applied for feature extraction and data fusion in many application domains including but not limited to chemistry, neuroscience, social network analysis and computer vision \cite{acar2009,kolda2008}. Prada et al. \cite{prada2012} have previously used a three-way analysis of SHM data for damage detection and feature selection. However, this work was only studied to achieve damage detection, not damage localization and estimation. In \cite{khoa2015}, Khoa et al. proposed the use of a tensor analysis approach for damage identification in SHM through the use of CANonical DECOMPosition (CANDECOMP) using PARAllel FACtors (PARAFAC) analysis (CP) decomposition. The CP decomposition has been shown to achieve fast convergence through the Alternating Least Squares algorithm \cite{rabanser2017introduction}, and is the tensor decomposition approach that will be used in this paper.


\section{Background Theory}

\subsection{Support Vector Machines and Anomaly Detection}
Consider a set of $m\in\mathbb{N}$ training data enumerated as follows:
\begin{align}
    (\bm{x}_1,\bm{y}_1),...,(\bm{x}_m,\bm{y}_m) \in \mathcal{X} \times {\pm 1}
\end{align}

where each $\bm{x}_i$ is from a non empty set $\mathcal{X}$ known as the \textit{domain}, and the $\bm{y}_i$ data points are known as the \textit{target} variables, which can take values of either -1, or 1. In binary classification the aim is to learn a \textit{decision function}, $f: \mathcal{X} \to {\pm 1}$. However in the case of anomaly detection which is the typical setting in SHM, we usually only have access to $\mathcal{X}$, and so the problem often becomes repurposed to finding an $f$ which takes value +1 in a particular region in space, and -1 elsewhere. Mathematically we solve what is known as the one-class support vector machine (SVM) problem, described by Scholkopf et al. \cite{scholkopf2000support} as follows:

\begin{align}
    \min_{\bm{w}\in F, \xi \in \mathbb{R}^m, \rho \in \mathbb{R}} \quad  \frac{1}{2} || \bm{w} ||^2 + \frac{1}{\nu m} \sum_{i=0}^m \bm{\xi}_i- \rho \\
    \text{subject to: } (\bm{w} \cdot \Phi(\bm{x}_i)) \geq \rho - \xi_i, \quad \xi_i \geq 0
\end{align}

where $\xi_i\in\mathbb{R}$ refers to the $i$-th slack variable, $\Phi: \mathcal{X} \to F $, represents a mapping of the $\bm{x}$ variable to a different feature space, $\bm{w}\in\mathbb{R}^{|F|}$ refers to the weights, $F$ is some dot product space in the image of $\Phi$, and the hyper-parameter $\nu \in (0,1)$, assists in characterising the solution in two ways: it is an upper bound on the fraction of allowed outliers, and is a lower bound on the number of support vectors \cite{scholkopf2002support,scholkopf2000support}. Through this optimisation problem, the SVM approach can be considered to be a \textit{maximum-margin classifier}, in that the distance from the $y_i = 1$, and $y_i = -1$ training pairs to the SVM boundary are maximised. Due to this property, the SVM boundary is known to be robust to new examples. 

It is also typical for SVMs to employ the use of a kernel, due to the \textit{kernel trick}, since it allows the dot product of two training domain examples in the projected space, to be computed efficiently through the evaluation of kernel function which corresponds to the particular projection ($\Phi$) function. That is, $k(\bm{x}_i,\bm{x}_j) = \langle \Phi(\bm{x}_i),\Phi(\bm{x}_j) \rangle.$ In this paper, the kernel used in all SVM calculations is the \textit{radial basis function} (RBF) kernel: 

\begin{equation}
   k(\bm{x}_i,\bm{x}_j)= \text{exp}(-\gamma || \bm{x}_i - \bm{x}_j ||^2),
\end{equation}

where $\gamma = \frac{1}{2\sigma^2} \in \mathbb{R}$, refers to the length-scale of the RBF kernel. Due to its formulation, and corresponding length-scale measure the RBF kernel has a good interpretation as a similarity metric (since the RBF kernel value decreases exponentially the further away the two points $x_i$, and $\bm{x}_j$ are, and the $\gamma$ variable controls this rate of decrease). With the one-class SVM used in this paper we aim to tune two hyper-parameters: $\nu$ and $\gamma$.




\subsection{Tensor Analysis for SHM Data}
\label{section:tensor}

\begin{figure}[!h]
	\centering
    \includegraphics[width=0.5\textwidth]{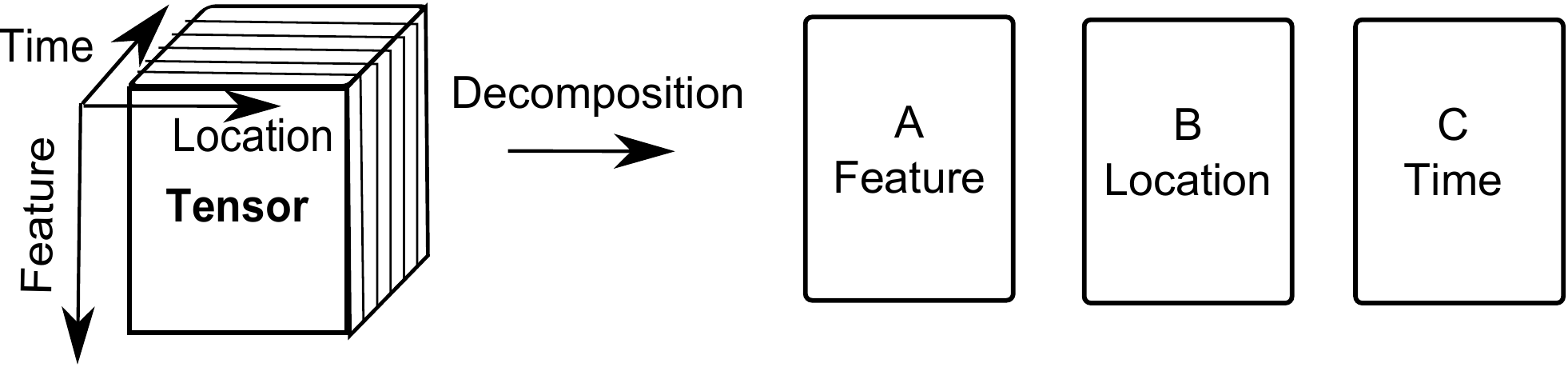}
	\caption{Typical tensor data in SHM.}
	\label{fig:tensor}
\end{figure}

In SHM applications, there are usually many sensors at different locations used to measure the vibration signals over time. In this case, the incoming data can be represented as a three-way tensor with dimensions of $(feature \times location \times time)$, as described in Figure \ref{fig:tensor}. The label, \textit{Feature}, refers to the information extracted from the raw signals in the time domain (for example, features from a frequency domain). \textit{Location} represents the relative sensor positioning, and \textit{time} refers to data snapshots at different time stamps. \cite{Cheema2016_SHM}

Two typical approaches for tensor decomposition are the CANonical DECOMPosition (CANDECOMP) using PARAllel FACtors (PARAFAC) analysis (CP) decomposition and the Tucker decomposition \cite{kolda2009}. However, due to the `core tensor form' of the Tucker decomposition, which is difficult to use and interpret, the CP form is used in this work.

The CP decomposition of a tensor factorizes the tensor as a sum of a finite number of rank-one tensors. In the case of a three-way tensor $\mathcal{X} \in \mathbb{R}^{I \times J \times K}$, it may be expressed as,
\begin{equation} \label{eqn:CP_decomp}
\mathcal{X} = \sum_{r=1}^R\lambda_r \bm{A}_{:r} \circ \bm{B}_{:r} \circ \bm{C}_{:r} + \mathcal{\bm{E}},
\end{equation}
where $R\in\mathbb{N}$ is the latent factor, $\bm{A}_{:r}$, $\bm{B}_{:r}$ and $\bm{C}_{:r}$ are $r$-th columns of component matrices $A \in \mathbb{R}^{I \times R}$, $B \in \mathbb{R}^{J \times R}$ and $C \in \mathbb{R}^{K \times R}$, and $\lambda\in\mathbb{R}^R$ is the weight vector so that the columns of $\bm{A}$, $\bm{B}$, $\bm{C}$ are normalized to length one. The symbol `$\circ$' represents a vector outer product, and $\mathcal{\bm{E}}$ is a three-way tensor containing the residuals.
It is shown in element-wise as
\begin{equation}
x_{ijk} = \sum_{r=1}^R\lambda_r a_{ir}b_{jr}c_{kr} + e_{ijk}.
\end{equation}
It can also be written in term of the $k$-th frontal slice of $\mathcal{X}$:
\begin{equation}
\label{equa:slideCP}
\bm{X}_k = \bm{AD}_k\bm{B}^{\intercal} + \bm{E}_k,
\end{equation}
where $\bm{D}_k$ is a diagonal matrix represented by $\text{diag}(\lambda_r \bm{C}_{k:})$, with $\bm{C}_{k:}$ means the `$k$-th row of matrix $\bm{C}$'. In order to obtain the matrices $\bm{A}, \bm{B},$ and $\bm{C}$, an algorithm known as \textit{alternating least squares} (ALS) is often used. In order to perform this algorithm, the matrix $\bm{A}$ (for example) is randomly initialised, and then updates are consecutively made to $\bm{A,B}$ and $\bm{C}$ as shown in Equation \ref{eqn:als}. \cite{rabanser2017introduction}

\begin{align} \label{eqn:als}
    \bm{A} \leftarrow \arg\min_{\bm{A}}  \| \bm{X}_{(1)} - \bm{A}(\bm{C} \odot \bm{B})^{\intercal} \| \nonumber \\
    \bm{B} \leftarrow \arg\min_{\bm{B}} \| \bm{X}_{(2)} - \bm{B}(\bm{C} \odot \bm{A})^{\intercal} \| \\
    \bm{C} \leftarrow \arg\min_{\bm{C}} \| \bm{X}_{(3)} - \bm{C}(\bm{B} \odot \bm{A})^{\intercal} \| \nonumber
\end{align}

where the symbol refers to the Khatri-Rao Product, which can be expressed as a column-wise Kronecker product.

\begin{align*}
    \bm{A} \odot \bm{B} = [\bm{\mathrm{a}}_1 \otimes \bm{\mathrm{b}}_1 \bm{\mathrm{a}}_2 \otimes \bm{\mathrm{b}}_2 \ldots \bm{\mathrm{a}}_k \otimes \bm{\mathrm{b}}_k]
\end{align*}

where for clarity we define: $\bm{A}\in \mathbb{R}^{I\times K}$ and $\bm{B}\in \mathbb{R}^{J\times K}$, resulting,  $ \bm{A} \odot \bm{B} \in \mathbb{R}^{I\times J K }$, and where $\bm{X}_{(i)}$ refers to the unfolding of tensor $\mathcal{X}$ in mode $i$.

The second benefit of using a CP decomposition approach lies in its flexibility to be used for online learning. In particular, Zhou et al. \cite{zhou2016accelerating} have proposed a technique known as \textit{onlineCP}, which allows the arrival of new data to be placed in $\bm{C}$-space. In this way once the SVM has been trained on $\bm{C}$ data, there is no need to re-update the underlying model in time, all that needs to be done is place each new data point into $\bm{C}$-space. This procedure can be performed by considering that,

\begin{align} \label{eqn:incremental_tensor}
     \bm{C} &= \arg\min_{\bm{C}} \| \bm{X}_{(3)} - \bm{C}(\bm{B} \odot \bm{A})^{\intercal} \| \nonumber \nonumber \\ 
       &= \arg\min_{\bm{C}} 
       \left\| 
      \begin{bmatrix}
      \bm{X}_{\text{old}{(3)}} \\
      \bm{X}_{\text{new}{(3)}} 
      \end{bmatrix} 
      -       
      \begin{bmatrix}
      \bm{C}^{(1)} \\
      \bm{C}^{(2)}
      \end{bmatrix}
      (\bm{B} \odot \bm{A})^{\intercal}
      \right\| \\
       &= \arg\min_{\bm{C}} 
       \left\| 
      \begin{bmatrix}
      \bm{X}_{\text{old}{(3)}} - \bm{C}^{(1)}(\bm{B} \odot \bm{A})^{\intercal}  \\
      \bm{X}_{\text{new}{(3)}} 
      -   \bm{C}^{(2)}(\bm{B} \odot \bm{A})^{\intercal}
      \end{bmatrix}
      \right\| \nonumber
\end{align}

Here Zhou et al. make the observation that $\bm{C}^{(1)}$ clearly is a minimiser for the first row of Equation \ref{eqn:incremental_tensor}, and thus in order to minimise the second row, we require $\bm{X}_{\text{new}{(3)}} = \bm{C}^{(2)}(\bm{B} \odot \bm{A})^{\intercal}$. Thus the value of $\bm{C}^{(2)}$ may be estimated by taking the psuedo-inverse of the matrix $(\bm{B} \odot \bm{A})^{\intercal}$. This results in the incremental tensor update equation, which is used in this paper to simulate the arrival of new data inputs, which is shown in Equation \ref{eqn:incremental_final}.

\begin{equation} \label{eqn:incremental_final}
    \bm{C}_{\text{new}} = \bm{X}_{\text{new}{(3)}} ((\bm{B} \odot \bm{A})^{\intercal})^\dagger
\end{equation}

\section{Methodology}

\subsection{The $N$-DoF Lagrangian Aeroservoelastic Model}

The use of aeroservoelastic (ASE) models has been well documented in aeronautical literature since the 1990's. Both Noll and Baker et al. have noted that ASE models have increased in relevancy at a similar rate  as aircraft have increased in size, due to the inevitable interactions between aerodynamic forces and structural dynamics \cite{ASEsum,ASEsum2}. Noll further suggests that the increasing complexity of modern systems coupled with the highly flexible and light structures naturally necessitates ASE modelling \cite{ASEsum2}. Typical ASE modeling considers the equations of motion being split into inertial, damping, and stiffness matrix terms, where the stiffness and damping matrices consist of both: structural and aerodynamic components \cite{ASEHeimbaugh, ASE1, ASE2}. The wing itself is usually modeled as a flat plate with freedom in pitch and plunge (a pitch-plunge model) where the structural stiffness is modeled by restraining springs in each degree of freedom. Initially the equations of motion are expressed as a set of second order ordinary differential equations, but are then usually discretised into a state space time domain model when running simulations \cite{ASEHeimbaugh, ASE3}. The typical ASE equation can be expressed as in Equation \ref{eqn:ASE_typical}.

\begin{equation} \label{eqn:ASE_typical}
    \left[
    \bm{M}\ddot{\bm{x}}+
    (\bm{C}_1-\frac{\rho c V}{4}\bm{C}_2)\dot{\bm{x}}
    +
    (\bm{K}_1-\frac{\rho V^2}{2}\bm{K}_2)x
    \right]
    = \bm{F}
\end{equation}

where $\bm{x}\in \mathbb{R}^N$ are the $N$ state variables, $\bm{M,C_1,C_2,K_1,K_2}\in\mathbb{R}^{N\times N}$ are the inertial, structural and aerodynamic damping, structural and aerodynamic terms, $c$ is the chord length of airfoil, $\rho$ is damping, $V$ is the velocity, and the system is driven by an external force $\bm{F}\in \mathbb{R}^N$. Usually the servo dynamics are included into these matrix equations so that this equation is not just aeroelastic, but aero\textit{servo}elastic. \cite{ASE3} Alternatively Pitt et. al. show how a basic $N$-DoF model can be expanded to include the influence of the servo. The servo (actuator) commonly used is a hydraulic piston with a connected control rod \cite{ASE1,wright2003use}, and as such will be used in this paper. It is assumed that the actuator is inertia-less given that its mass is not significant relative to the order of magnitude of frequencies concerned with the wing model. The model used will be based on that developed previously by Wright et al. \cite{wright2003use} It is widely commented that this solution is a good approximation of low damped modes \cite{ASEHeimbaugh, ASE1, ASEsum}.

\begin{figure}[!h]
\centering
\subfloat[Locations of sensor placement on the $N$-DoF ASE wing model. This version of the discretisation was extensively studied in this paper.]{\includegraphics[width=0.75\textwidth]{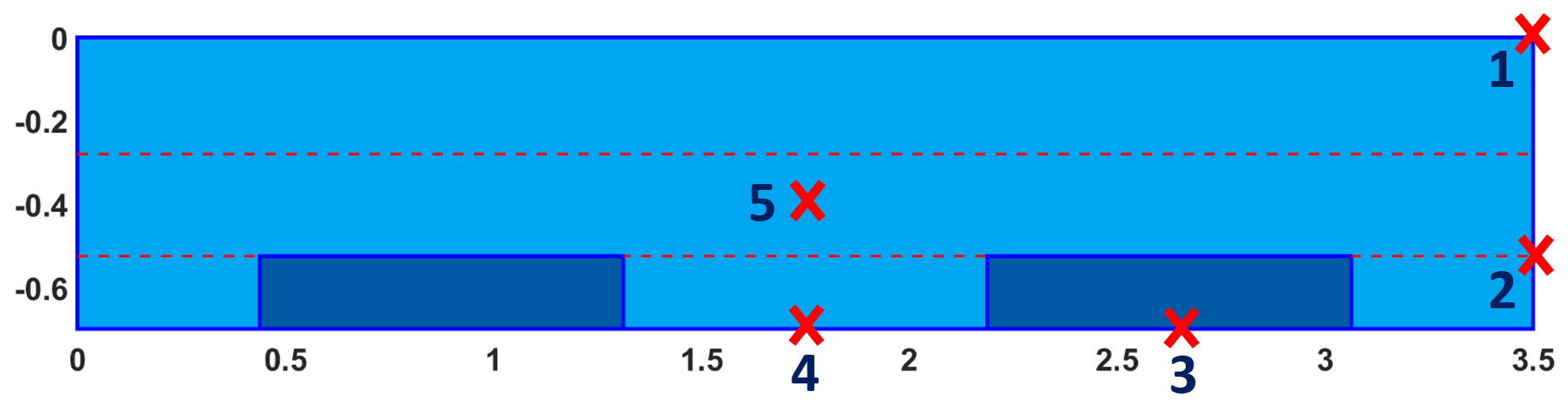}\label{fig:wing_without_act}}\\
\subfloat[Discretisation convention for the $N$-DoF ASE wing model. Random control surface discretisations are shown.]{\includegraphics[width=0.75\textwidth]{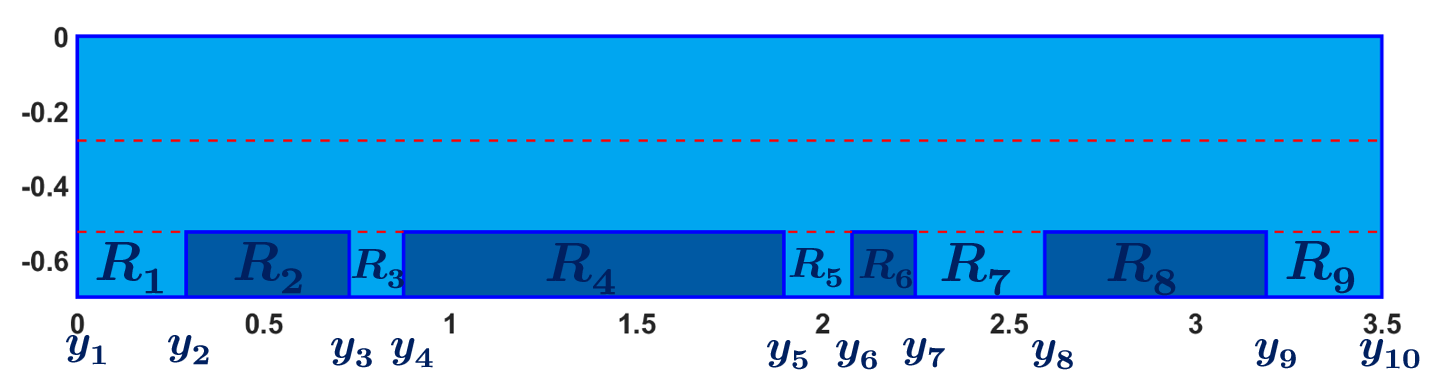}\label{fig:wing_with_act}}
\caption{Numbering and conventions used for the $N$-DoF ASE wing model.\label{fig:ASE_convention}}
\label{fig:wing_act}
\end{figure}

The ASE model developed in this paper will be derived from an energy-based Lagrangian perspective. Structurally the wing considered is 2D and rectangular, and will consist of a set of $M$ control surfaces, of which each of the $M$ control surfaces will be attached to one hydraulic piston each, resulting in $M$ independent hydraulic pistons in total. The aerodynamics will be implemented by considering Theodorsen's strip theory, with a quasi-steady aerodynamic assumption, where the quasi-steady assumption is defined as the limiting effect of oscillatory aerodynamics as the system frequency moves towards zero, which is consistent with the definition of Hancock et al. \cite{hancock1985teaching}. This way, it is possible to capture the complexity of the state space in much more simpler form than one would otherwise obtain via pure kinetics, or through FE-model coupling. We begin by considering Equation \ref{eqn:Lag_basic} and \ref{eqn:Lag_basic_2}:

\begin{align} \label{eqn:Lag_basic}
L = T - V
\end{align}

\begin{align} \label{eqn:Lag_basic_2}
\frac{d}{d t}
 \left(\frac{\partial T}{\partial \dot{q}_i}\right) + \frac{\partial T}{\partial q_i} + \frac{\partial V}{\partial q_i} + \frac{\partial F}{\partial \dot{q}_i} = Q_i
\end{align}

where $L$ denotes the Lagrangian, $T$ refers to kinetic energy, $V$ refers to potential energy, $q_i$ is the $i$-th general co-ordinate, where $i \in \{1,...,N | N\in\mathbb{N}\}$, $F$ refers to the dissipative force of the system \cite{minguzzi2015rayleigh}, $Q_i=\frac{\partial \delta W}{\partial \delta\dot{q}_i}$ represents the generalised force acting on the system, and the dot notation refers to the time derivative. The states, $\bm{q}$ which we consider for this system are the wing bending, $\gamma$, the wing twist $\theta$, and the control surface deflections $\beta_j$, where $j \in \{1,...,M | M = N - 2\}$,, where $q\in \mathbb{R}^N$. That is $\bm{q} = \left[\gamma,\theta,\beta_1,\beta_2, ... , \beta_M\right]$. In this way there are $N$ degrees of freedom for this model, $M$ of which are $\beta$, corresponding to a total of $M$ total control surfaces. When the actuator model is introduced the $q$ vector will become of size $2M + 1$ since there is only one actuator per control surface. The actuator state variable will be represented by the $P_J$, and will refer to the pressure differential over the entire actuator. The $y$-positions on the wing where the control surface begins and ends are given by a sequence of ordered numbers, $y_i\in\mathbb{R}$, where $y_{i+1} \geq y_{i}$. This is clarified in Figure \ref{fig:ASE_convention}.

\begin{figure}[!h]
	\centering
    \includegraphics[width=0.8\textwidth]{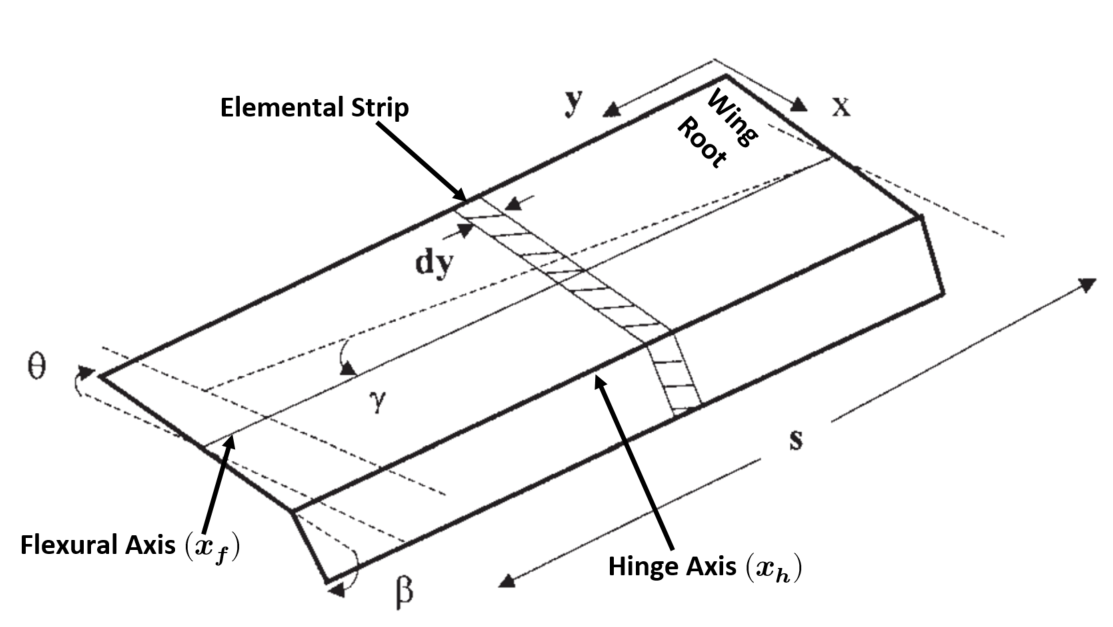}
	\caption{Co-ordinates and labelling conventions on the ASE wing system. Based on Wright et al. \cite{wright2003use} \label{fig:ASE_lablled}}
\end{figure}

As a result of this formulation, the coordinate of any point on top of the wing will be given by 

\begin{align} \label{eqn:z_normal}
   z = y\gamma + (x-x_f)\theta + \sum_{j=1}^{M}(x-x_h)\beta_j\bm{1}_{\mathcal{R}_{\text{2j}}}(y)\bm{1}_{x>x_h}(x),
\end{align}

such that,

\begin{align} \label{eqn:regions}
    \mathcal{R} &= \{\left[y_{i},y_{i+1}\right] \mid i= 1,...,2M-1,\text{ where } \mu\left(\footnotesize{\bigcup}_i \mathcal{R}_i\right) = s \},
\end{align}

so that $\mathcal{R}_i$ refers to the closed interval $[y_{i},y_{i+1}]$. Here, the $y\in\mathbb{R}_{+}$ and $x\in\mathbb{R}_{+}$ variables refer to the co-ordinates of the wing as clarified in Figure \ref{fig:ASE_lablled}, $x_f\in\mathbb{R}_{+}$, $x_h\in\mathbb{R}_{+}$ refer the the flexural and control surface hinge axes,  $\bm{1}_{\mathcal{X}} : \mathcal{X}\to \{0,1\}$ is the indicator function, $\mathcal{R}$ refers to the number scheme used to segment the wing into a variety of regions as Figure \ref{fig:ASE_convention} clarifies, and $\mu$ is the standard Lebesgue measure. Note that Equation \ref{eqn:regions} implies the following equation:

\begin{equation}
    \bigcap_{i} \mathcal{R}_i = \varnothing
\end{equation}

Now, differentiating Equation \ref{eqn:z_normal} with respect to time gives:

\begin{align} \label{eqn:z_diff}
       \dot{z} = y\dot{\gamma} + (x-x_f)\dot{\theta} + \sum_{j=1}^{M}(x-x_h)\dot{\beta}_j\bm{1}_{\mathcal{R}_{2j}}(y)\bm{1}_{x>x_h}(x).
\end{align}

Equation \ref{eqn:z_diff} allows us to find the Lagrangian of the aeroservoelastic system, since $T = 1/2 M \dot{z}^2$, where $M = \int \int_{\mathcal{S}} \rho dm$, and $\rho$ is the material density of the wing, and where $\mathcal{S}$ refers to the entire wing surface (note that the wing is approximated as a 2D surface rather than a 3D volume). In order to do so we make use of the generalised formulation for an arbitrary sequence of elements, $a_p$, where $p = 1,..,P$, which is shown in Equation \ref{eqn:exp_terms}. 

\begin{align} \label{eqn:exp_terms}
\left( \sum_{p=1}^P a_p \right)^2 = \sum_{p=1}^P a_p^2 + 2 \sum_{s=1}^{P}\sum_{t=1}^{s-1} a_t a_s ,
\end{align}

With the expression at hand we find that the kinetic energy of this $N$-DoF system may be expressed by Equation \ref{eqn:big_fkn_integral}.

\begin{multline} \label{eqn:big_fkn_integral}
 T = \frac{1}{2} \int_0^s \int_0^c \left(y\dot{\gamma}\right)^2 dm +
 \frac{1}{2} \int_0^s \int_0^c \left((x-x_f)\dot{\theta}\right)^2 dm + 
 \frac{1}{2} \int_0^s \int_0^c \sum_{j=1}^{M}\left((x-x_h)\dot{\beta}_j\bm{1}_{\mathcal{R}_{2j}}(y)\bm{1}_{x>x_h}(x)\right)^2 dm + \\
 \cancelto{0}{\int_0^s \int_0^c \sum_{s=1}^{M}\sum_{t=1}^{s-1}(x-x_h)^2\dot{\beta}_s\bm{1}_{\mathcal{R}_{2s}}(y)\bm{1}_{x>x_h}(x)\dot{\beta}_t\bm{1}_{\mathcal{R}_{{2t}}}(y)\bm{1}_{x>x_h}(x) dm} +
 \int_0^s \int_0^c y\dot{\gamma}(x-x_f)\dot{\theta} dm + \\
 \int_0^s \int_0^c y\dot{\gamma}\sum_{j=1}^{M}(x-x_h)\dot{\beta}_j\bm{1}_{\mathcal{R}_{2j}}(y)\bm{1}_{x>x_h}(x) dm +
 \int_0^s \int_0^c (x-x_f)\dot{\theta}\sum_{j=1}^{M}(x-x_h)\dot{\beta}_j\bm{1}_{\mathcal{R}_{2j}}(y)\bm{1}_{x>x_h}(x) dm
\end{multline}

Where each of the area moment of inertia terms can be collected from Equation \ref{eqn:big_fkn_integral} as follows:

\begin{align} \label{eqn:I_integrals}
    I_{\gamma \gamma} &= \int_{0}^{s} \int_{0}^{c} y^2 dm\\
    I_{\theta \theta} &= \int_{0}^{s} \int_{0}^{c} (x-x_f)^2 dm \\
   I_{\beta_j \beta_j} &= \int_{\mathcal{R}_{2j}} \int_{x_h}^{c} (x-x_h)^2 dm = t\rho\mu(\mathcal{R}_{2j})\int_{x_h}^{c} (x-x_h)^2 dx\\
    I_{\gamma \theta} &= I_{\theta \gamma} = \int_{0}^{s} \int_{0}^{c} y(x-x_f) dm\\
    I_{\gamma \beta_j} &= I_{\beta_j \gamma} = \int_{\mathcal{R}_{2j}} \int_{x_h}^{c} y (x-x_h) dm \\
    I_{\theta \beta_j} &= I_{\beta_j \theta} = \int_{\mathcal{R}_{2j}} \int_{x_h}^{c} (x - x_f)(x-x_h) dm = t\rho\mu(\mathcal{R}_{2j})\int_{x_h}^{c} (x - x_f)(x-x_h) dx
\end{align}

where, $\mu$ is the Lebesgue measure. Since the $y$ integral is univariate, and the integrand is measurable (in particular Lebesgue measurable), the measure terms $\mu$ simply evaluate to the length of the interval. For example, $\mu({\mathcal{R}_{4}})=y_5 - y_4.$ Note that the lack of absolute value stems from the ordering of $y$, since $y_{i+1} \geq y_i$. In addition notice that a large part of the kinetic energy expression cancels out to zero, due to the presence of multiple indicator functions which cannot activate in unison. The system has been designed in this way due to the structural constraints developed previously (made clear in Equation \ref{eqn:regions}. Also from some of the expressions in the inertia integrals we make clear that the density, $\rho$ and thickness, $t$ can be removed from the mass differential $dm$, making clear our assumptions about constant thickness and constant density along the wing. However we opt to leave $dm$ in where possible from here onwards to highlight the general nature of this derivation. 

Although the generalised kinetic energy term has been derived, we still must derive the potential energy of the structure. However this cannot be derived from first principles, and so basic assumptions will need to be made. In particular the potential energy we consider is shown in Equation \ref{eqn:potential_energy},

\begin{equation} \label{eqn:potential_energy}
 V = \frac{1}{2}K_{\gamma\gamma}\gamma^2 + \frac{1}{2}K_{\theta\theta}\theta^2 + \frac{1}{2}\sum_{j=1}^M K_{{\beta\beta}_j} \beta_j^2,
\end{equation}

which assumes independence between all the state variables in terms of structural stiffness. We also assume that each stiffness term, $K$, can be obtained through the classic stiffness-frequency relationship: $\omega = \sqrt{\frac{K}{I}}$. That is we must specify the $\omega$ frequency terms for each state variable (which is system dependent), in order to back-calculate the corresponding stiffness terms, given that we know the $I$ expressions as a result of calculating the kinetic energy terms. 

Up until this point, we have briefly outlined the derivation for the kinetic energy and structural potential energy terms for use in the Lagrangian equation. However currently it is only an $N$-DoF elastic model, and not aeroelastic. Thus additional aerodynamics are required. The terms will be included via a generalised $N$-DoF 2D strip theory approach with quasi-steady aerodynamic assumptions. The strip theory equations for the incremental lift, wing moment, and hinge moment expressions are outlined in Equations \cref{eqn:dL,eqn:dM,eqn:dH}. These equations can be obtained from \cite{wright2008introduction}, but have been generalised here to consider the presence of multiple control surfaces.

\begin{align}
dL &= \frac{1}{2}\rho V^2 c dy \left[a\left(\theta +  \frac{\dot{\gamma}y}{V}\right) + a_c\sum_{j=1}^M\beta_j\bm{1}_{\mathcal{R}_{2j}}(y)  \right]\label{eqn:dL}\\
dM &= \frac{1}{2}\rho V^2 c^2 dy \left[ae\left(\theta + \frac{\dot{\gamma}y}{V} \right) + a_m\sum_{j=1}^M\beta_j\bm{1}_{\mathcal{R}_{2j}}(y) + M_{\dot{\theta}}\frac{\dot{\theta}c}{V}  \right]\label{eqn:dM}\\
dH &= \frac{1}{2}\rho V^2 c^2 dy \left[b_1\left(\theta +  \frac{\dot{\gamma}y}{V} \right) + b_2\sum_{j=1}^M\beta_j\bm{1}_{\mathcal{R}_{2j}}(y) + \frac{M_{\dot{\beta} }c}{V}\sum_{j=1}^M\dot{\beta}_j\bm{1}_{\mathcal{R}_{2j}}(y)  \right]\label{eqn:dH}
\end{align}

Moreover the incremental work on the entire aeroelastic system can be defined as in Equation \ref{eqn:inc_work}.

\begin{align} \label{eqn:inc_work}
    \delta W = -\int_0^s dL y \delta \gamma + \int_0^s dM \delta \theta + \int_0^s dH \sum_{j=1}^M \delta \beta_j \bm{1}_{\mathcal{R}_{2j}}(y)
\end{align}

This is needed in order to obtain the final generalised aerodynamics forces for the system through: $Q_{q_i} = \frac{\partial (\delta W)}{\partial (\delta q_i)}$ which can then be substituted into Equation \ref{eqn:Lag_basic_2}. 

In addition, the following Remark \ref{remark:measure} is developed to aid in simplifying Equations \cref{eqn:dL,eqn:dM,eqn:dH}. In particular it describes the way in which the $\sum_{j=1}^M\beta_j\bm{1}_{\mathcal{R}_{2j}}(y)$ terms can be broken down into simpler expressions. This is done to maintain the generality of developed equations, and for keeping expressions very succinct. However assuming the aforementioned ordering of $y$ points, and assuming that all the control surfaces are rectangular and lie along the hinge axis, then the reader can mentally replace every instance of $\mathcal{R}_{2j}(y)$ with $y_{2j+1} - y_{2j}$.

\begin{remark} \label{remark:measure} The integral of the sum of discrete control surface angles over the wing, $\int_0^L \sum_{j=1}^M \beta_j \bm{1}_{\mathcal{R}_{2j}}(y) dy$, can be expressed succinctly as a sum of Lebesgue measures. That is, \end{remark}

\begin{align*}
\int_0^L \sum_{j=1}^M \bm{1}_{\mathcal{R}_{2j}}(y) dy & = \sum_{j=1}^M \beta_j \int_0^L  \bm{1}_{\mathcal{R}_{2j}}(y) dy\\
&= \sum_{j=1}^M \beta_j \mu(\mathcal{R}_{2j}).
\end{align*}

\begin{figure*}[!h]
\centering
\subfloat[Pitch-plunge aeroelastic model without the actuator model.]{\includegraphics[width=0.45\textwidth]{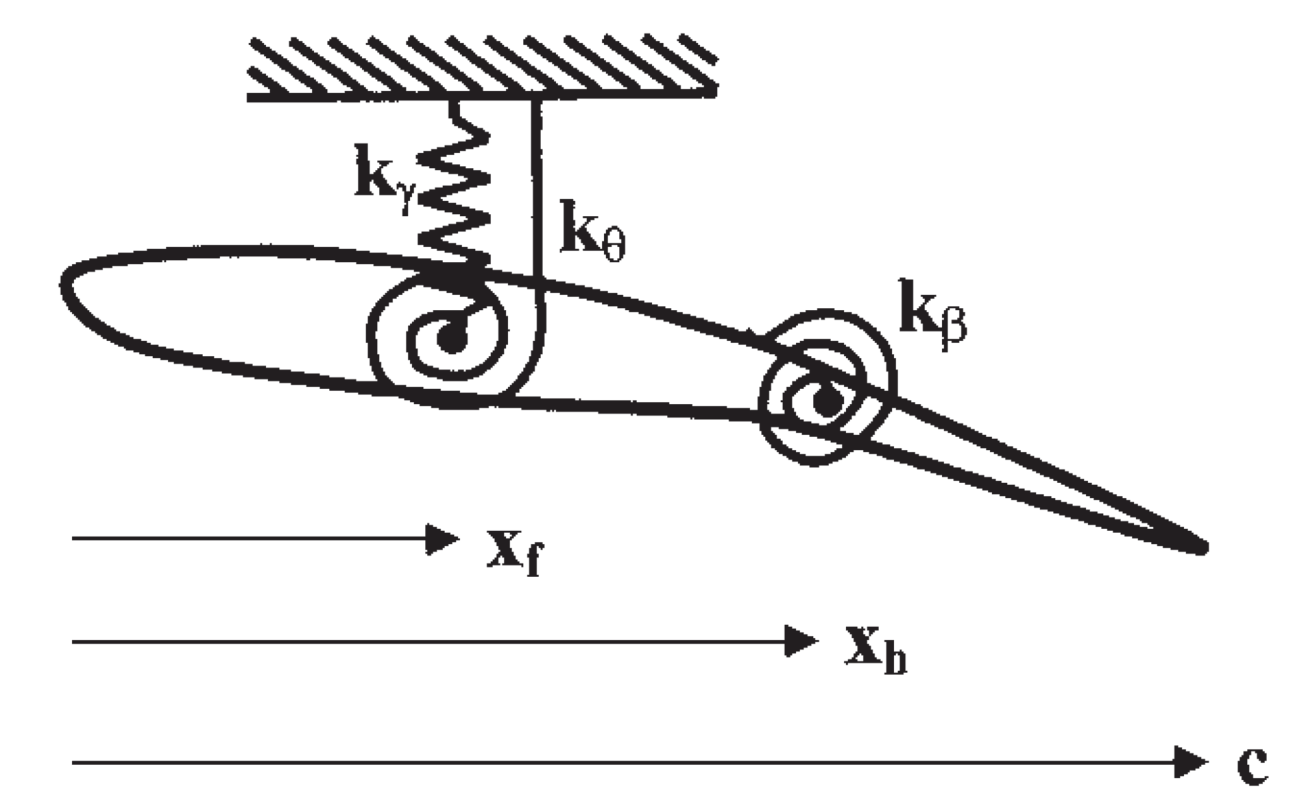}\label{fig:wing_without_act}}
\subfloat[Pitch-plunge aeroelastic model with the actuator model.]{\includegraphics[width=0.45\textwidth]{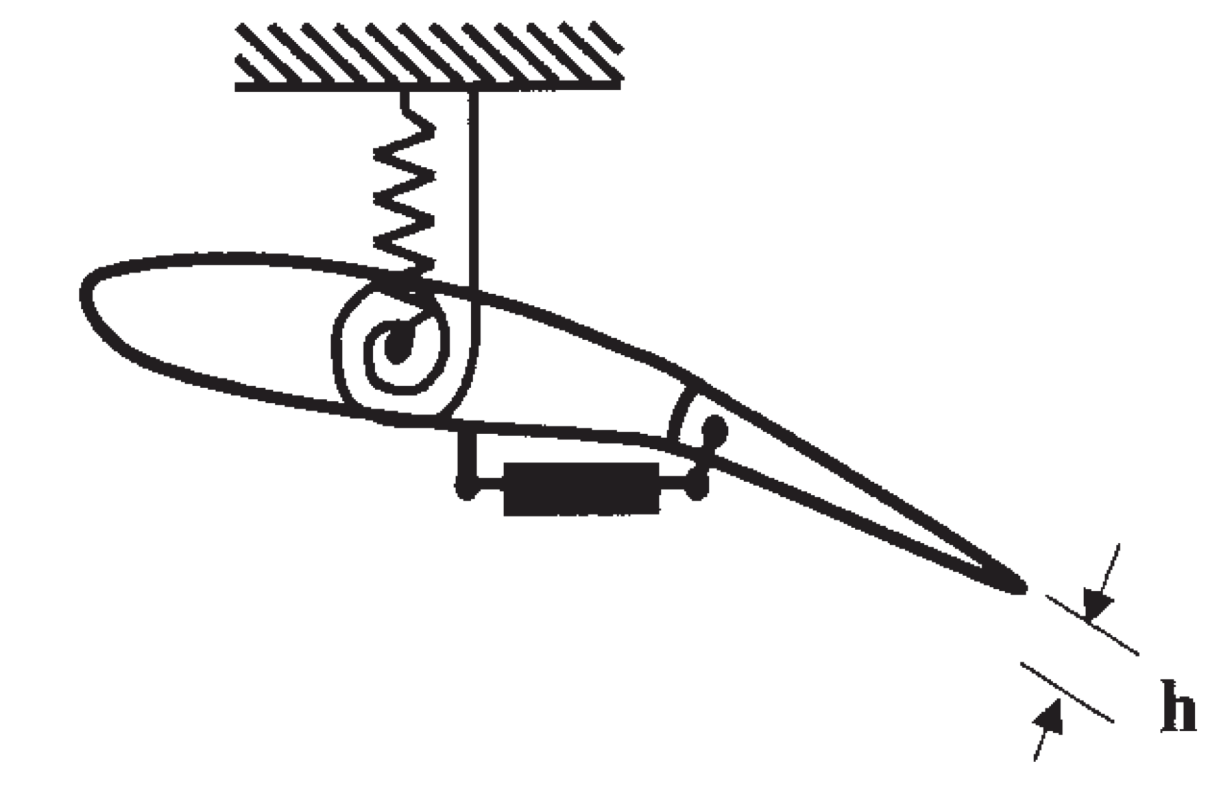}\label{fig:wing_with_act}}
\caption{Comparison of the wing aeroelastic wing model, with and without the presence of the linearised actuator model.\cite{wright2003use} \label{fig:pitchplunge}}
\label{fig:wing_act}
\end{figure*}

Lastly, the servo model used in this paper is based on one previously developed by Wright et al., which is linearised and massless. \cite{wright2003use} The linearisation simplifies the overall dynamics, and the massless nature is just an assumption stemming from the idea that the natural frequency of the piston will be extremely minor when compared to that of the entire wing. In this ASE model in this paper, there is one hydraulic actuator per control surface. Exactly how the actuator assembles onto the \textit{pitch-plunge} airfoil model is shown in Figure \ref{fig:pitchplunge}.

\begin{figure}[!h]
	\centering
    \includegraphics[width=0.6\textwidth]{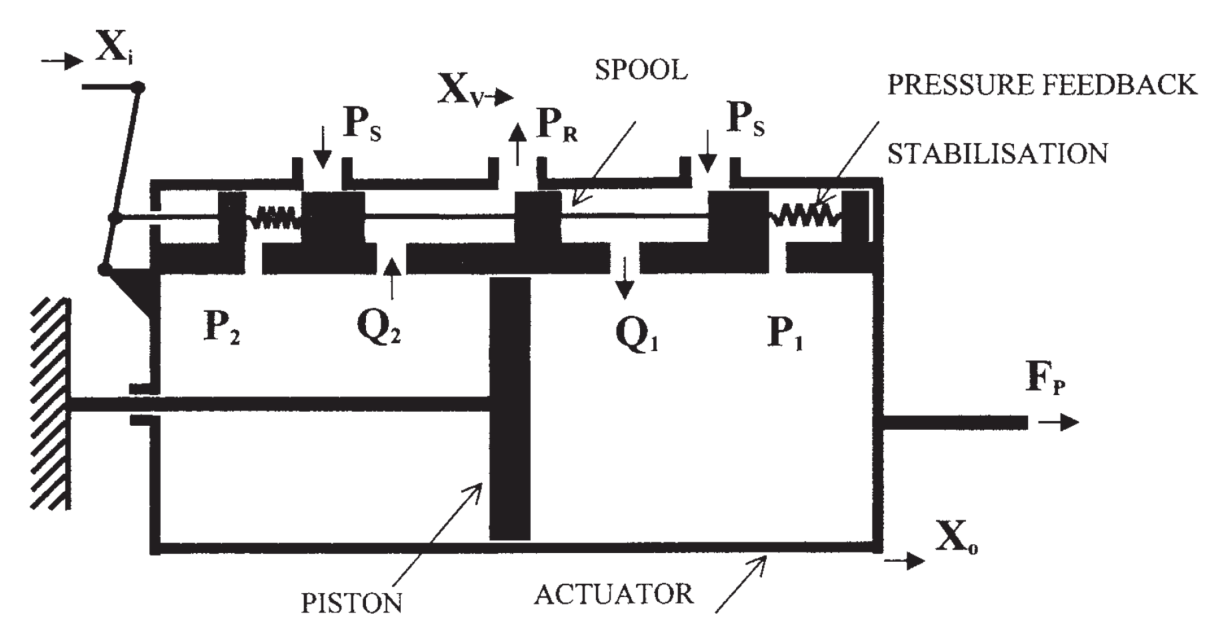}
	\caption{Diagram of actuator model used, from Wright et al.\cite{wright2003use} \label{fig:piston}}
\end{figure}

In particular, based on on Wright et al. \cite{wright2003use} we have the following equation for each control surface:

\begin{align}
    -h A_p \dot{\beta}_j + \frac{V_0}{4N}\dot{P}_{{J}_j}-h\mu K_V\sqrt{\frac{P_s}{2}}\beta_j + \frac{K_V A_F}{K_F}\sqrt{\frac{P_s}{2}}P_{{J}_j}=-h\mu K_V \sqrt{\frac{P_s}{2}} \beta_{C_j}
\end{align}

where $K_V$ is a valve flow constant, $P_R$ and $P_S$ are the return and supply pressures, $\beta_C$ is the commanded deflection angle for the control surface, $P_J$ is a state variable referring to the pressure differential over the entire actuator, $A_P$ is the cross sectional area of the piston, $A_F$ is the area of the feedback spring, $V_0$ is the rate of change of the piston oil volume, $\mu$ is the lever arm ratio, $N$ is the bulk modulus of the oil, and $h$ is the distance orthogonal distance offset from the control surface to the piston, which enables kinematic relationships between the piston and control surface to be derived. Moreover in order to implement this model, consistent with Wright et al. we assume that $K_{{\beta\beta}_j} \rightarrow 0$. We also assume (as mentioned earlier) that the inertia of the piston in comparison with the overall wing dynamics is negligible. Details on how all the variables work together in the piston is shown in Figure \ref{fig:piston}. Thus, by combining all the previous information into a large set of linear equations we arrive at Equation \ref{eqn:fkn_huuuuge}, which is placed in Appendix A due to its size. All additional constants and parameters required for Equation \ref{eqn:fkn_huuuuge} are available in Wright et al. \cite{wright2008introduction}.

\subsection{Signal Processing for the $N$-DoF Aeroservoelastic Model}

\begin{figure}[!h]
	\centering
    \includegraphics[width=1.0\textwidth]{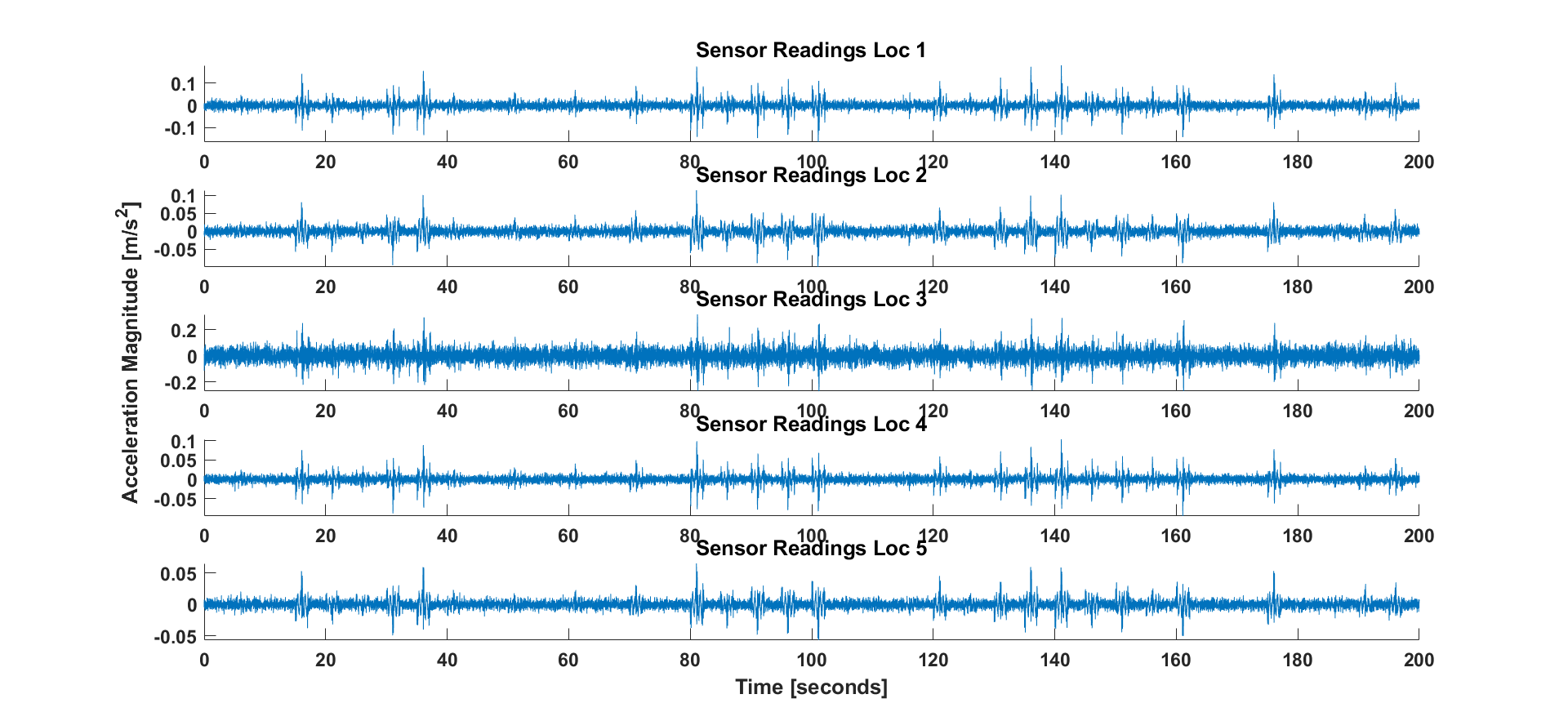}
	\caption{Example of simulated accelerometer sensor readings available from the ASE model.}
	\label{fig:sensor_readings}
\end{figure}

Now that the $N$-DoF ASE system has been formalised, it is possible to use it to extract artificial sensor data for the purpose of damage detection. In particular, the dynamical system described in Equation \ref{eqn:fkn_huuuuge} can be rearranged into a state space form, which can then be used to simulate acceleration data from a sensor. The locations and numbering of the sensors used in this study have already been outlined in Figure \ref{fig:ASE_convention}. Example acceleration readings are shown in Figure \ref{fig:sensor_readings}. The acceleration readings can be obtained anywhere on top of the wing model by differentiating Equation \ref{eqn:z_diff}, and Gaussian white noise has been added to the readings in order to simulate the presence of atmospheric turbulence. Moreover, the magnitude of the response from sensor three is far larger than any other sensor, which can be explained when noting that sensor three is located on the control surface, and so has the most direct response to commanded input angles. Note that although the signals gathered are in the time domain, they are cleaned for noise and passed through a fast Fourier transform in order to transform it to frequency domain data. This is because the feature space for this paper is in the frequency domain,

\begin{figure}[!h]
\centering
\subfloat[A \textit{neat} ordering of possible commanded input angles, considering a range of all possible angles (large and small). ]{\includegraphics[width=0.45\textwidth]{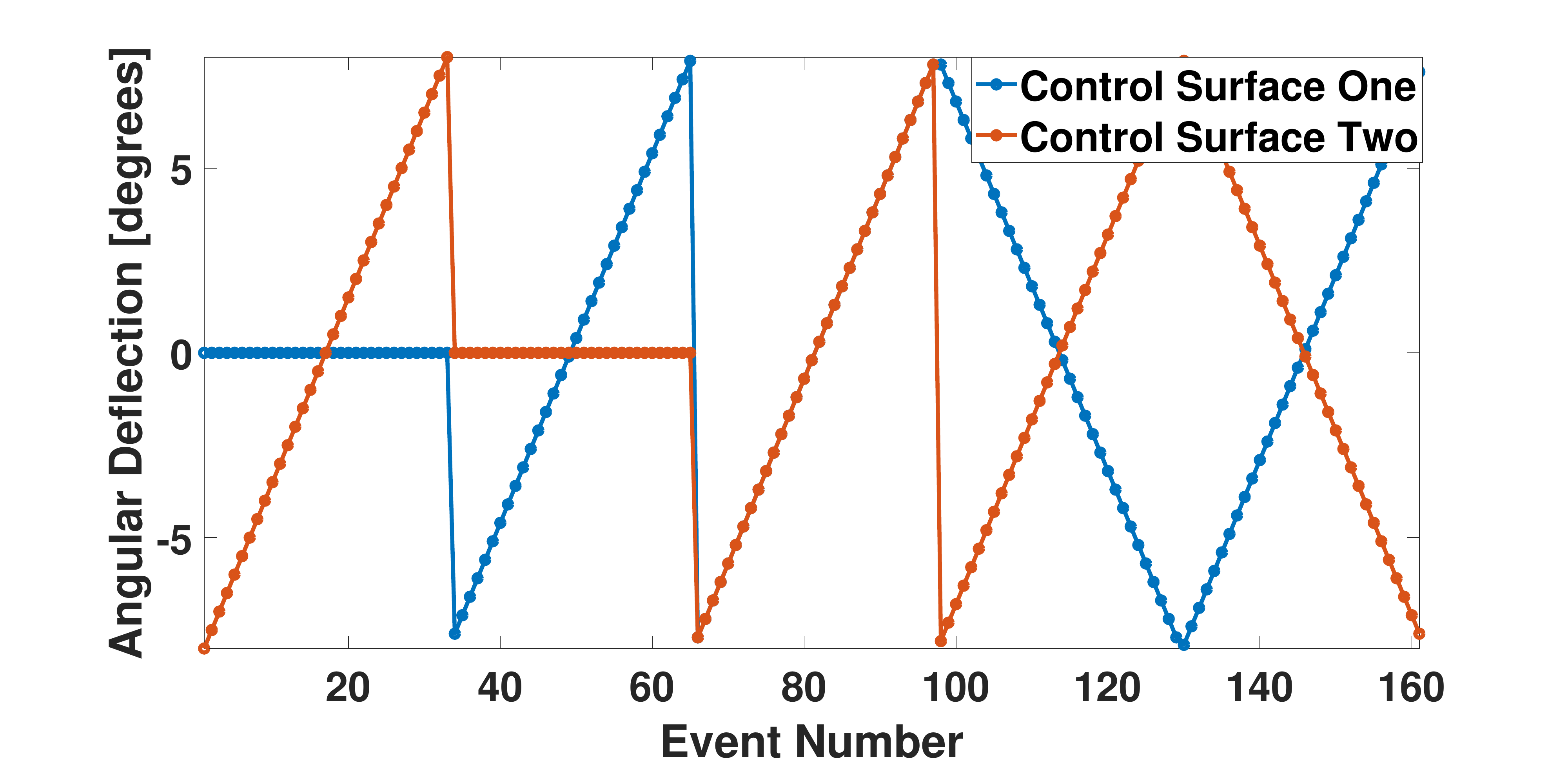}\label{fig:input_defl1}}
\subfloat[An ordering of possible input angles, only considering large angles (greater than 5 degrees), and generated by LHS sampling. ]{\includegraphics[width=0.45\textwidth]{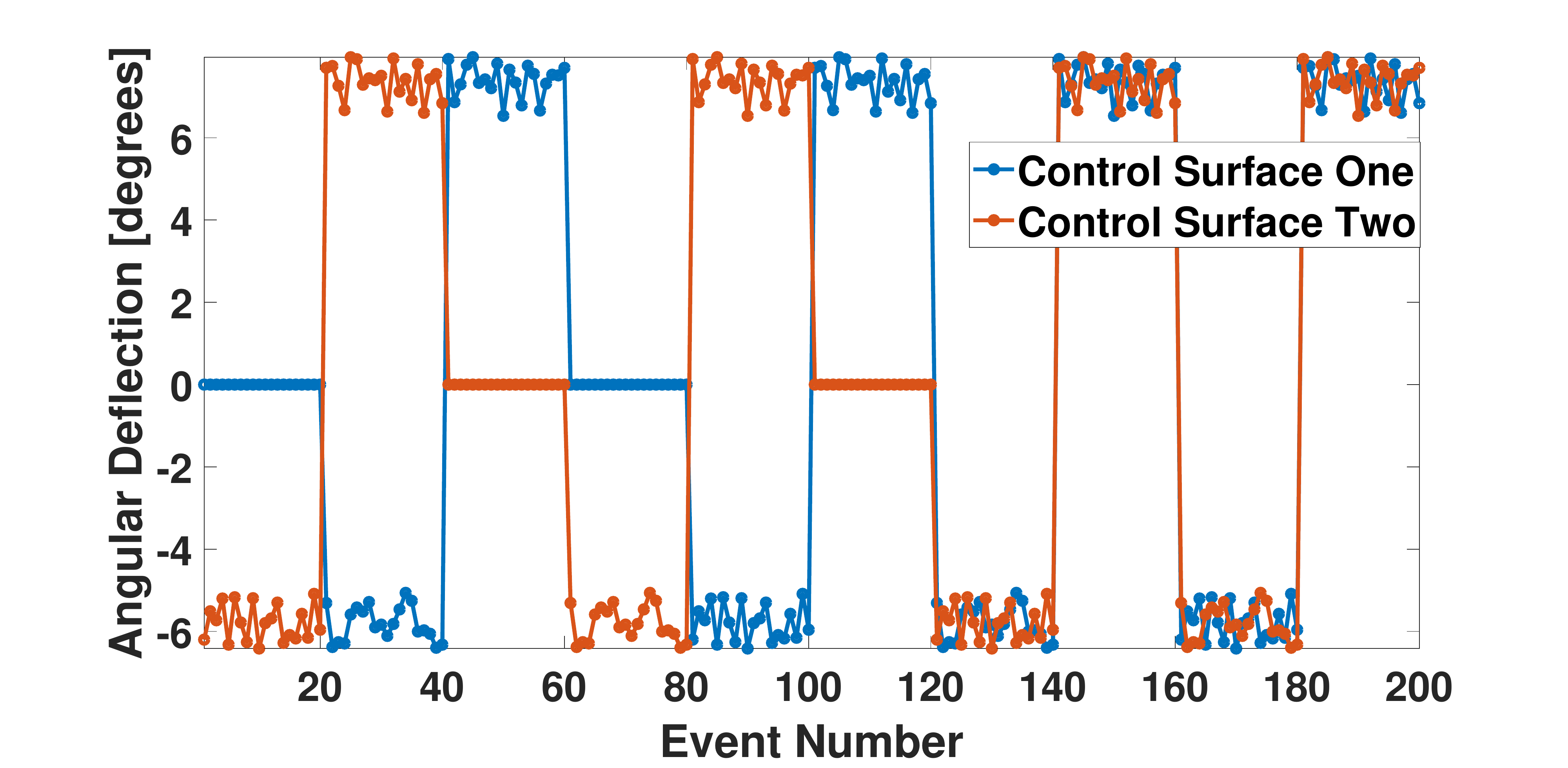}\label{fig:input_defl2}}
\caption{Comparing different methods to simulate input data for the $N$-DoF wing ASE system.}
\label{fig:input_defl}
\end{figure}

Each of the events shown in Figure \ref{fig:sensor_readings} correspond to a permutations of commanded input deflections of the control surfaces. There are many possible ways to generate this input permutation in order to generate the data, and some of those used in this paper are shown in Figure \ref{fig:input_defl}. More specifically Figure \ref{fig:input_defl1} shows a case where input angles were generated equi-spaced over a regular grid across all angle size ranging from 8 to -8. However Figure \ref{fig:input_defl2} shows another case study where the control surfaces were made to deflect by large amounts ($\pm[5,8]$), and the angles were selected by a Latin Hypercube Sampling (LHS) methodology. More information, on the use of LHS for the selection of points for aerospace structural system is shown in \cite{cheema2016experimental}. Note also that emphasis is placed on the word \textit{permute} rather than \textit{combination} because the ordering of input angles is asymmetric. The control surface(s) closer to the wing root will invariable give smaller magnitude responses than those closer to the wing tip, due to the effect of cantilever bending.

Lastly, although using the $N$-DoF aeroservoelastic system it is possible to simulate healthy, the purpose of this paper is to perform damage detection, and so it is necessary to also simulate the presence damage. There are many places where damage may occur on the wing, but the most obvious for this system is via damage in the actuator. The primary reason why a hydraulic actuation system may perform outside of its range of expected behaviour is due to the presence of pressure leakage. There are numerous reasons why pressure leakage may occur in a hydraulic actuation system. These include but are not limited to: gasket head damage, shaft sealant failure, leakage occurring from damaging hoses, faulty pumps. ~\cite{chun2007investigation,silva1977reliability} Therefore in this paper the presence of damage has been simulated by reducing the values of the supply pressure $P_s$, and two sources of internal spring stiffness, $k_f$, and $k_o$ in the actuator by 30\% each.

\section{Results and Discussion}
Here we present results demonstrating the use of the CP decomposition tensor analysis approach, and how it can be used in conjunction with ASE models for damage detection. 

\subsection{Comparing Dimensionality}

Healthy and unhealthy data were simulated for the $N$-DoF ASE model, and the one class SVM was trained on the healthy data. The data used for clustering was based on a tensor decomposition of the $\bm{C}$-space, and all new incoming testing data (both healthy and unhealthy) were pre-processed by using the approximation for $\bm{C}_{\text{new}}$ (Equation \ref{eqn:incremental_final}). 

\begin{figure}[H]
\centering
\subfloat[One-class support vector clustering on data in the $\bm{C}$-space.]{\includegraphics[width=0.50\textwidth]{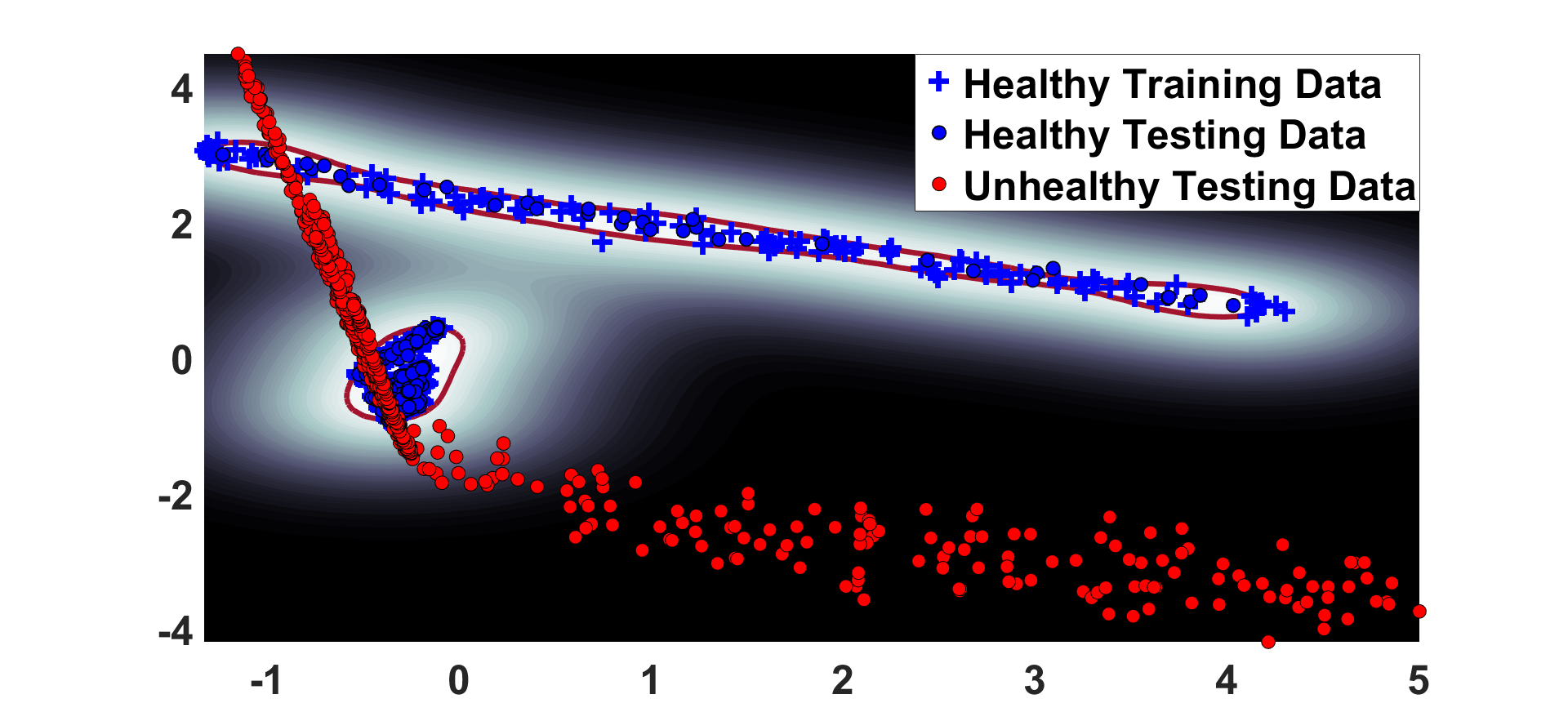}\label{fig:2d_C_space_1_a}}
\subfloat[Zoomed in version of Figure \ref{fig:2d_C_space_1_a}, showing the presence of clusters.]{\includegraphics[width=0.50\textwidth]{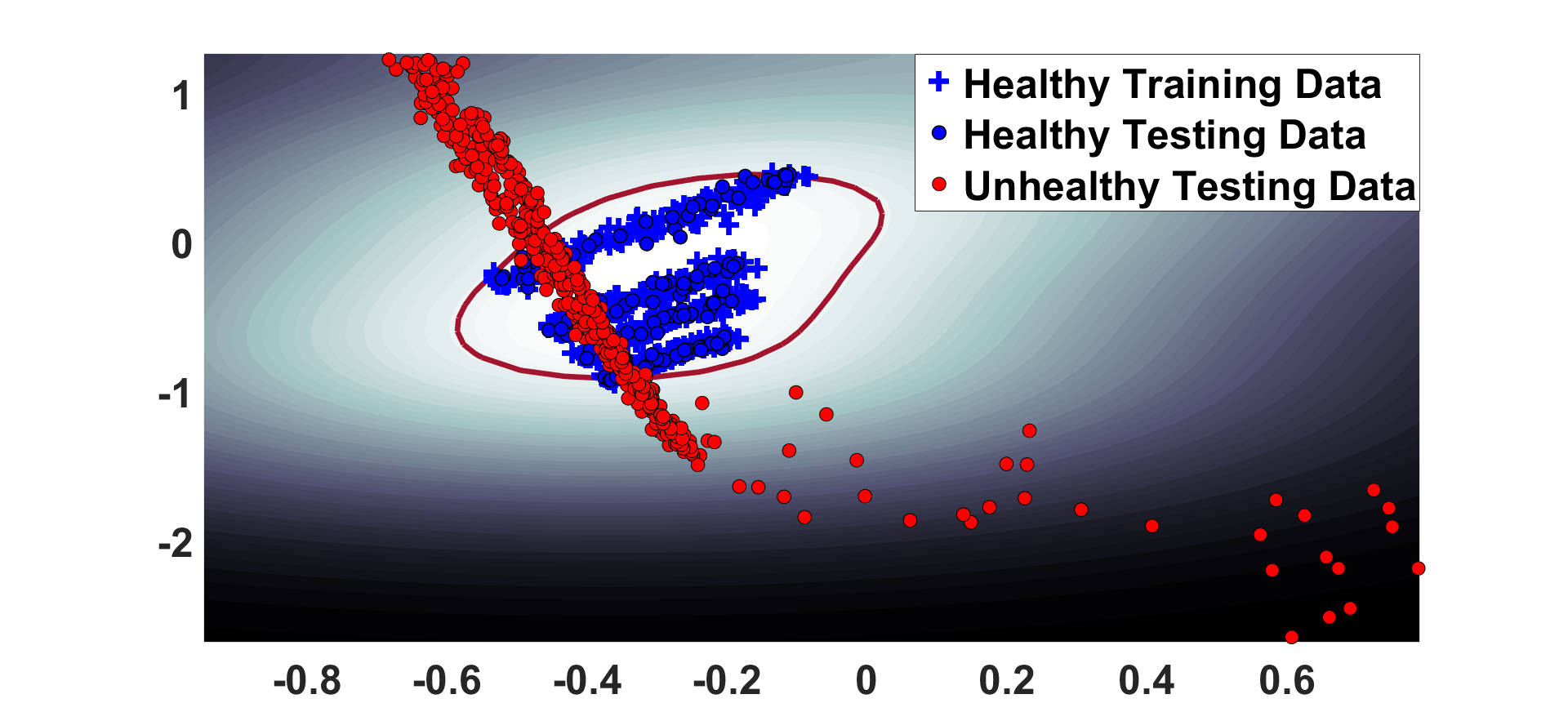}\label{fig:2d_C_space_1_b}}
\caption{Two dimensional representation of $C$-space data points used for damage detection.}
\label{fig:2d_C_space_1}
\end{figure}

From Figure \ref{fig:2d_C_space_1_a} it would appear that there are two prominent clusters in $\bm{C}$-space. However upon zooming in we observe that there are in fact several clusters present. Each of these clusters is strongly influenced by one of the senors on the wing model. In particular the large cluster towards the top of Figure \ref{fig:2d_C_space_1_a} corresponds more strongly to sensor three. This is because sensor three is attached to the control surface and so will be influenced the most due to commanded input angles, as opposed to the other sensors which will only be reacting to the commanded input angle through nonlinear interaction terms. Important to note is that the commanded input angles used to generate this data is that from Figure \ref{fig:input_defl1}, and so the input has a regular grid spacing, which is not entirely representative of random inputs that a conventional system may see. This issue will be explored in later sections.

\begin{figure}[!h]
	\centering
    \includegraphics[width=0.8\textwidth]{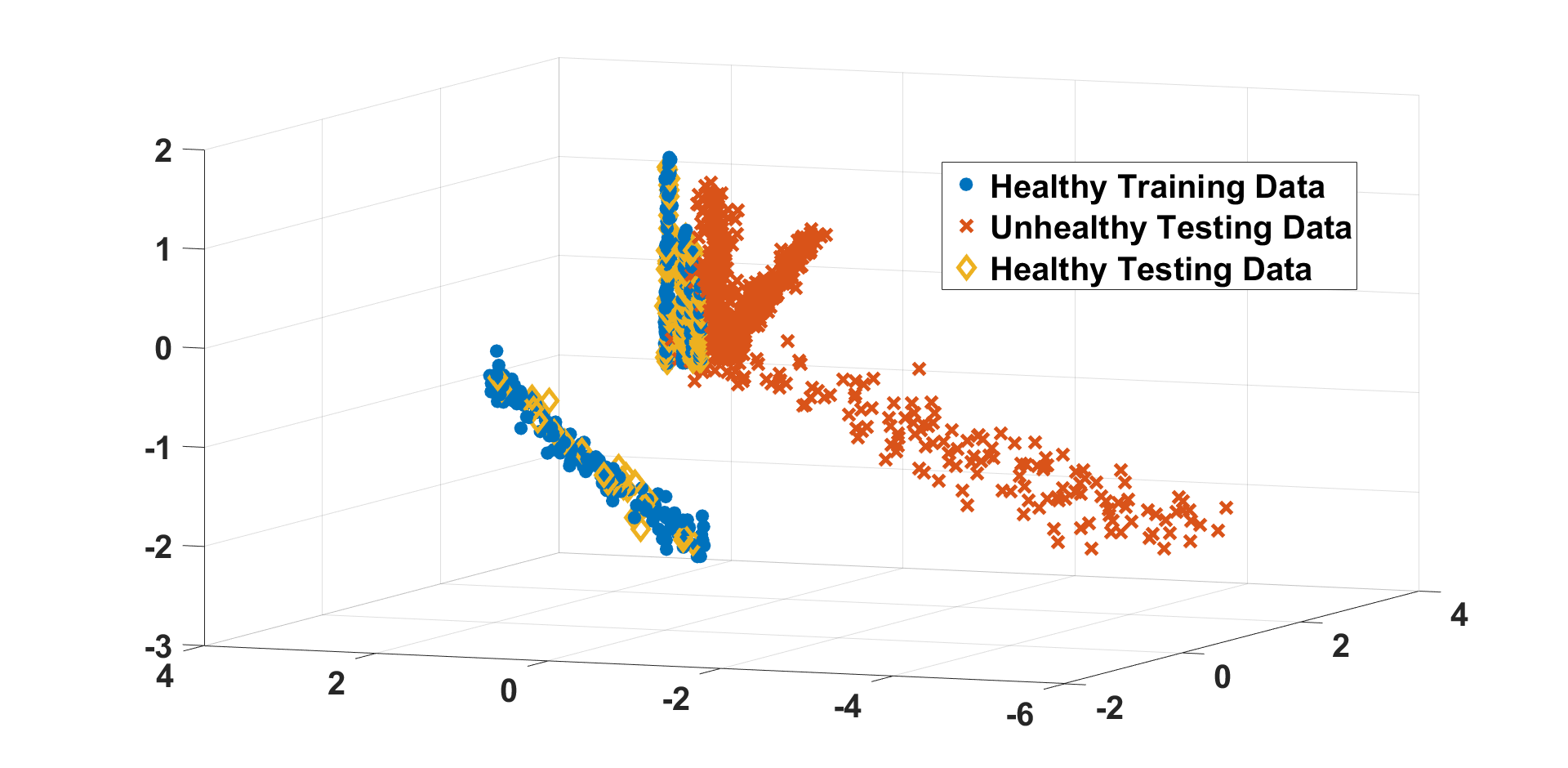}
	\caption{3D $\bm{C}$-space corresponding to the same data as shown in Figure \ref{fig:2d_C_space_1}.}
	\label{fig:3D}
\end{figure}

Also as seen from Figure \ref{fig:2d_C_space_1}, a lot of the unhealthy data passes over the healthy cluster which will naturally effect the final model scores in a negative way. There will be a much higher occurrence of false positives and false negatives. However it is possible to increase the accuracy of the score by working in a higher dimensional $\bm{C}$-space. In particular we can alter the $CP$ decomposition from $R=2$ up to $R=3$, resulting in a 3D projection space. The result of this is shown in Figure \ref{fig:3D}.

We may see from Figure \ref{fig:3D} that although the data in 2D space overlapped, and was inseparable for the one class SVM, in 3D space the data and clusters do become visibly separable, which is a result of how higher dimensional spaces tend to \textit{space-out} points. The amount of overall improvement to the underlying accuracy (among other metrics) and statistics for the one class SVM model is made clear in Table \ref{tab:tab1_3D}.

\begin{table}[!h]
\centering
\begin{tabular}{ll|ll}
\multicolumn{2}{c|}{\textbf{R=2}} & \multicolumn{2}{c}{\textbf{R=3}} \\ \hline
\multicolumn{1}{c}{546} & \multicolumn{1}{c|}{259} & \multicolumn{1}{c}{778} & \multicolumn{1}{c}{27} \\
\multicolumn{1}{c}{14} & \multicolumn{1}{c|}{148} & \multicolumn{1}{c}{7} & \multicolumn{1}{c}{155} \\ \hline
$\nu$: & 0.05 & $\nu$: & 0.03 \\
g: & 1.2 & g: & 1.2 \\ \hline
$F_1$: & 0.52 & $F_1$: & 0.90
\end{tabular}
\caption{Confusion matrices, hyper-parameters, and scores for the one class SVM model in two and three dimension, given the angle inputs specified in Figure \ref{fig:input_defl1}. Damage data is considered as \textit{positive}, and undamaged data as \textit{negative} for the confusion matrix. \label{tab:tab1_3D}}
\end{table}

As can be seen the $F_1$ scores increased form 0.52 up to 0.90, hinting towards a better model to use in practice. Moreover the confusion matrices have also changed, with the $R=3$ model having an order of magnitude less false positives, and half the amount of false negatives. These results imply that working in higher dimensional space is desirable, and in general for data which is not separable, it has been shown in vast amounts of kernel literature that projecting into a higher dimensional feature space does tend to make data more amenable to separation for clustering. \cite{hofmann2006review} However care must be taken here, since we are projecting into a higher dimensional space using the tensor decomposition, and so it is possible to project up to arbitrarily large spaces, which will separate all points from one another. Moreover it is possible to introduce artificial sources of noise in these higher dimensions. \cite{bro2003new} Therefore where possible it is preferred to look for methods which try to keep the dimensionality of the $\bm{C}$-space as low as possible in order to avoid such problems, and also to increase interpretability, since the 2D space is much simpler for future SHM engineers to interpret and evaluate.

\subsection{Comparing Input Angle Magnitude}

In the previous section, focus was placed on analysing how the difference in dimensionality of the $\bm{C}$-space may affect the results of a one class SVM classifier. Moreover the commanded angles input into the system were done so in a regular repeating grid structure. Here we explore how the differences in the input of the angles can affect the SVM model, and the $\bm{C}$-space. 

\begin{figure}[!h]
	\centering
    \includegraphics[width=0.8\textwidth]{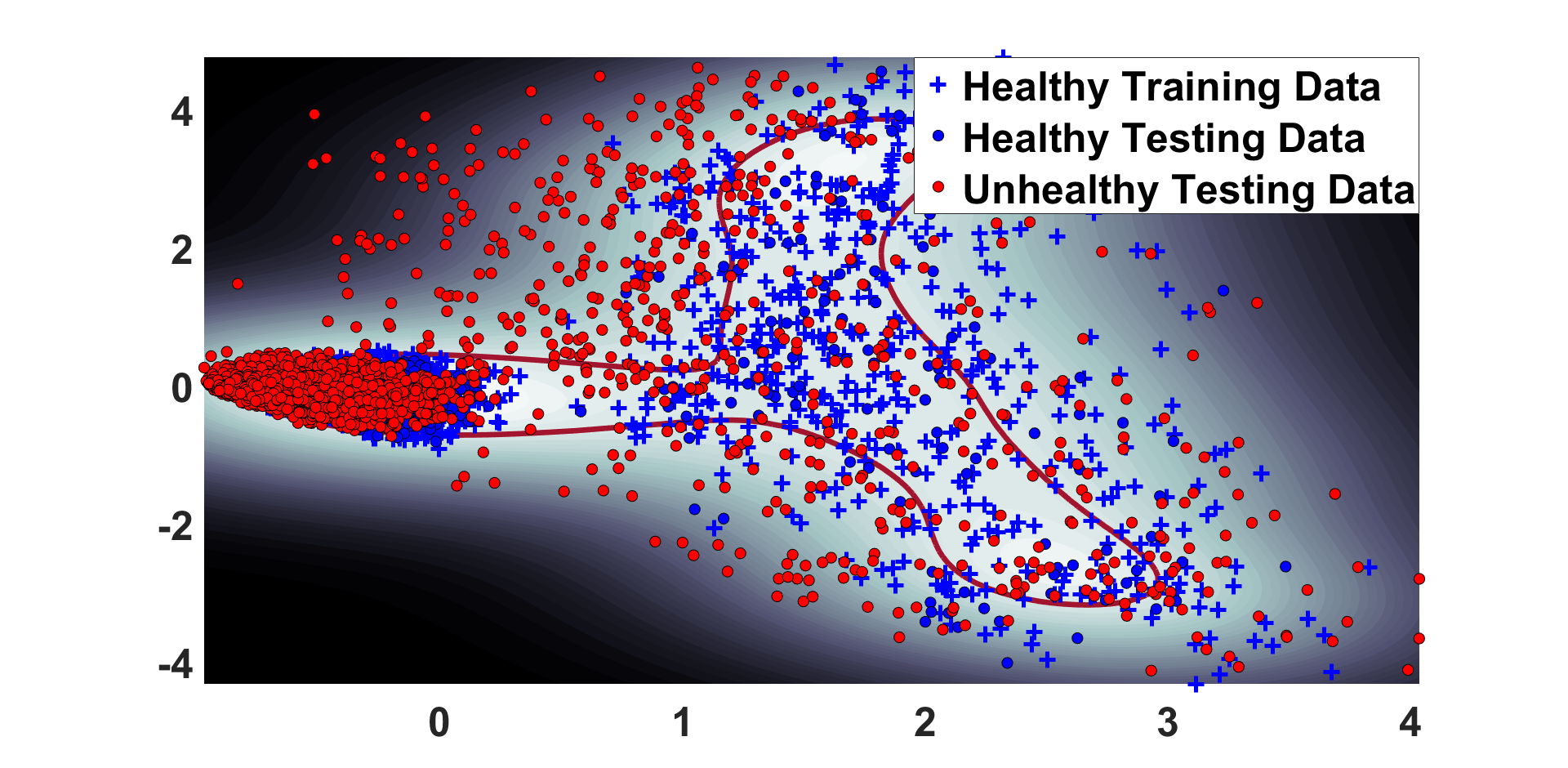}
	\caption{$\bm{C}$-space as a result of inputs being from an LHS, where each commanded input angle, $\beta_{C_1}$,$\beta_{C_2}\in [-8^{\circ},8^{\circ}]$.}
	\label{fig:LHS_all}
\end{figure}

The first difference here is, instead of inputting the angles as a regular grid (as shown in Figure \ref{fig:input_defl1}), the inputs now come from a Latin Hypercube sample (LHS). This reflects a more realistic setting for the possibilities and combinations for potential angle inputs, since there is increase randomness in the inputs, which will decrease the ability for the CP decomposition to \textit{cleanly} decompose the input tensor. The effect of now applying an LHS to the input angles is made clear in Figure \ref{fig:LHS_all}.

It can be seen that all the clusters which were previously observable when the inputs were ordered have now been mixed, and are now difficult to differentiate. Moreover we notice that the unhealthy data is now extremely difficult to separate from the healthy training set, there is a lot of overlap between the data points. This naturally increases the quantity of false positives and false negatives, and also works to lower the overall $F_1$-score. This is made clear in the first column of Table \ref{tab:tab2_LHS}. In addition, it can be seen that the one class SVM boundary also tries to be \textit{overly} inclusive and encompass all the healthy classes simultaneously owing to their close proximity to one another. This can make the training of SVMs difficult for use in aerospace structures since it appears that the $\bm{C}$-space for aerospace structures tends to naturally form clusters, and it is natural for SVMs to try and encompass everything. It is in fact very difficult to try and train a model that focuses individually on each structure, which is a negative aspect of using one class SVMs for aerospace structural modeling. This appears to be a different phenomenon as compared to previous literature which focused on civil structures, since civil structures tend to vibrate as one \textit{collective} mass, and so different accelerometers will be experiencing similar readings. \cite{Cheema2016_SHM,khoa2015} However in the aerospace field there are many active surfaces, which can move independently from one another, and also wings in particular experience a lot of bending and rotation during flight, especially closer towards the wing tips, which results in a $\bm{C}$-space that tends to naturally cluster.    

In order to initially try and enforce these clusters to be amenable for one class SVM analysis, artificial negative data was generated in accordance with Cheema et al. \cite{Cheema2016_SHM}. The is done so that the underlying one class problem can be treated equivalently as a two class problem with tighter boundaries. Cheema et al. have demonstrated this working well on civil structures by using kernel density estimates (KDE), Gaussian mixture models (GMM), and by local outlier factor (LOF) analysis. However it was reported by Cheema et al. that the KDE tends to work better due to its non-parametric nature, and so it was opted for here. It is non-parametric in that it estimates the probability density of data points via a summation of kernel functions ~\cite{bishop2013pattern}. Equation \ref{eqn:KDE} represents the generic form of an equi-weighted kernel density estimator~\cite{bishop2013pattern}:

\begin{equation} \label{eqn:KDE}
p(\mathbf{x}) = \frac{1}{N}\sum_{n=1}^N \frac{1}{h^D} k\left( \frac{\mathbf{x}-\mathbf{x_n}}{h} \right).
\end{equation}

In Equation \ref{eqn:KDE}, $p(\mathbf{x})$ is the estimated density at the point $\mathbf{x}\in\mathcal{X}\subseteq\mathbb{R}^N$, $h\in\mathbb{R}$ is the \textit{bandwidth} which acts as a spacing factor, $N\in\mathbb{N}$ is the total number of points in the distribution with its division ultimately acting as a normaliser, $D\in\mathbb{N}$ is the dimensionality of the feature space, and $k:\mathcal{X}\times\mathcal{X}\to\mathbb{R}$ is the kernel function. In this paper, a Gaussian kernel is used for negative sample generation, and its mathematical form is shown in Equation \ref{eqn:KDE_Gaussian}:

\begin{equation} \label{eqn:KDE_Gaussian}
p(\mathbf{x}) = \frac{1}{N}\sum_{n=1}^N \frac{1}{(2\pi h^2)^{1/2}} \text{exp}\left\lbrace -\frac{||\mathbf{x}-\mathbf{x_n}||}{2h^2} \right\rbrace,
\end{equation}
where $h$ now can be interpreted as representing the standard deviation of the Gaussian components. 

\begin{figure}[!h]
\centering
\subfloat[Points generated as a result of KDE .]{\includegraphics[width=0.55\textwidth]{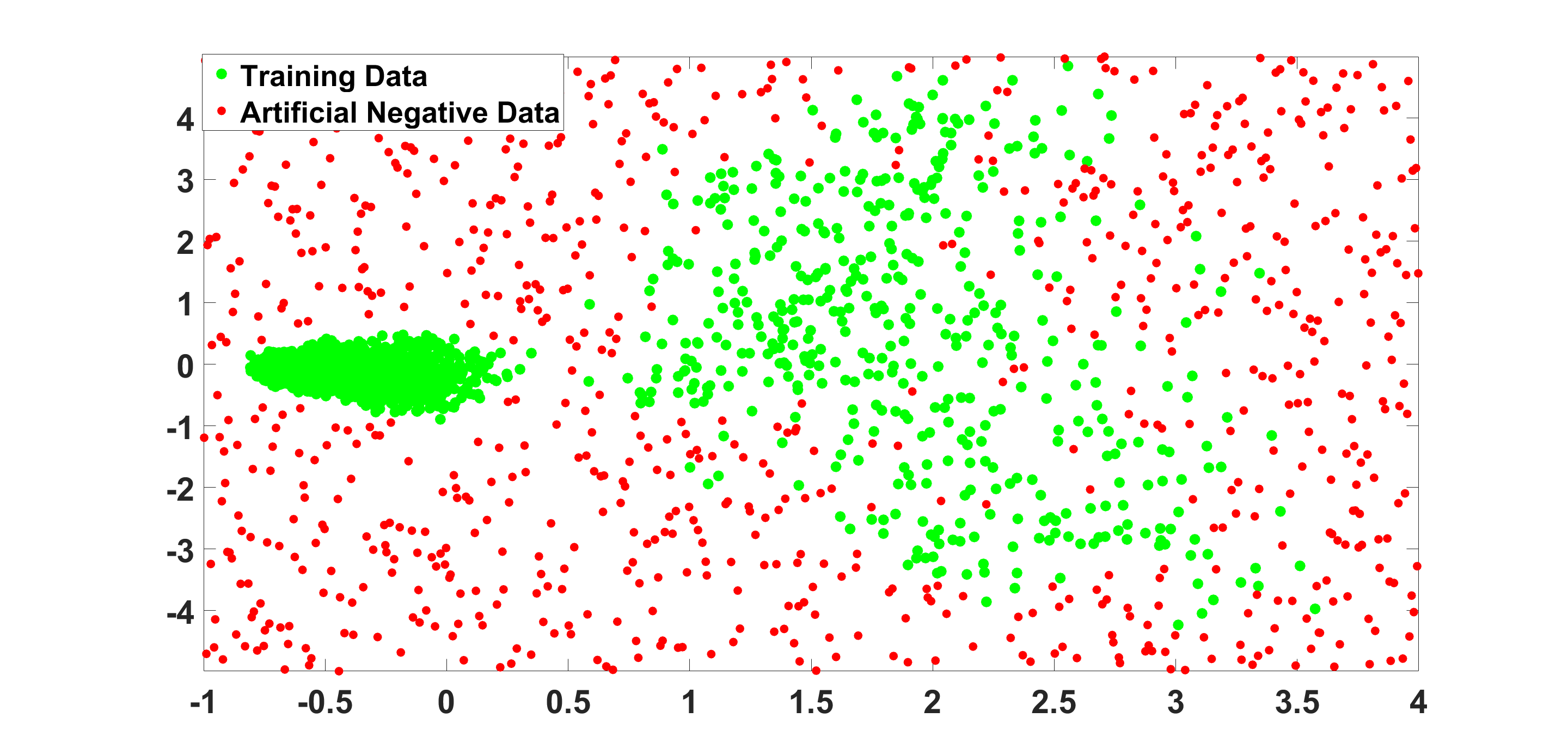}\label{fig:KDE_a}}\hspace{-2cm}
\subfloat[The density generated as a result f KDE.]{\includegraphics[width=0.55\textwidth]{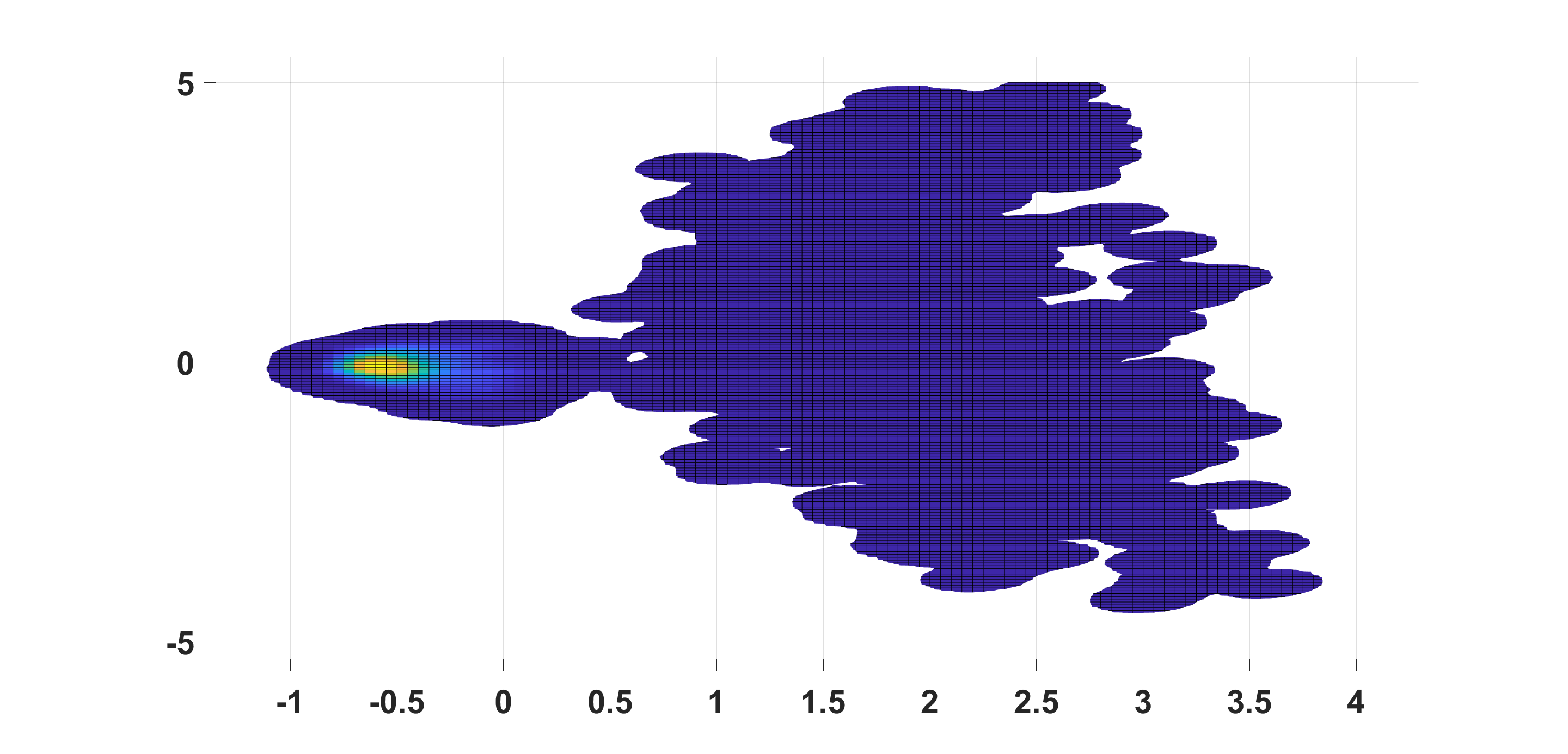}\label{fig:KDE_b}}
\caption{How a KDE approach can generate artificial negative data in order to transform the one class problem, into a two class problem.}
\label{fig:KDE}
\end{figure}

The results of using KDE to estimate a probability density over the training data in order to generate artificial outliers is shown in Figure \ref{fig:KDE}.

From Figure \ref{fig:KDE_a} it can be seen that KDE has successfully generated artificial negative training data in $\bm{C}$-space, and it has successfully done so in between the boundaries of the two clusters (where the more separate, larger cluster correlates strongly to sensor three, and every other sensor correlates strongly to the smaller, more dense cluster). However there are still many short comings of applying this \textit{fix} to this problem, which can all be seen by observing Figure \ref{fig:KDE_b}. Firstly there is still an overlap of density between the two clusters. Even though they visually appear separate, choosing the optimal length-scale for the Gaussian kernel in the KDE is difficult because smaller length scales will give highly overfit boundaries as the density peaks will be sharper, meaning that the KDE estimate will loose a lot of smoothness, which will in turn effect the boundary around the \textit{sparse} points towards the outside of the cluster associated with sensor three. However for slightly larger length scales, the two clusters will being to overlap, meaning that the boundary will still encompass the two clusters. Thus in this case, because of the closeness of the two clusters, and the sparsity of the cluster associated with sensor three, it becomes a difficult balancing act to find a \textit{useful} density fit. Alternate to a KDE estimation a GMM could be used, however once again looking at the density map of Figure \ref{fig:KDE_b}, the non-parametric densities exhibit behaviour which are not completely Gaussian. The smaller cluster appears to have a fatter tail, and the larger cluster follows a strange, non-Gaussian shape. Thus it would appear that the adding randomness in the input data to simulate a more realistic scenario makes the underlying classification much more difficult.

\begin{figure}[!h]
\centering
\subfloat[$\bm{C}$-space showing the data points, if the larger angles are considered.\label{fig:LHS_large_a}]{\includegraphics[width=0.45\textwidth]{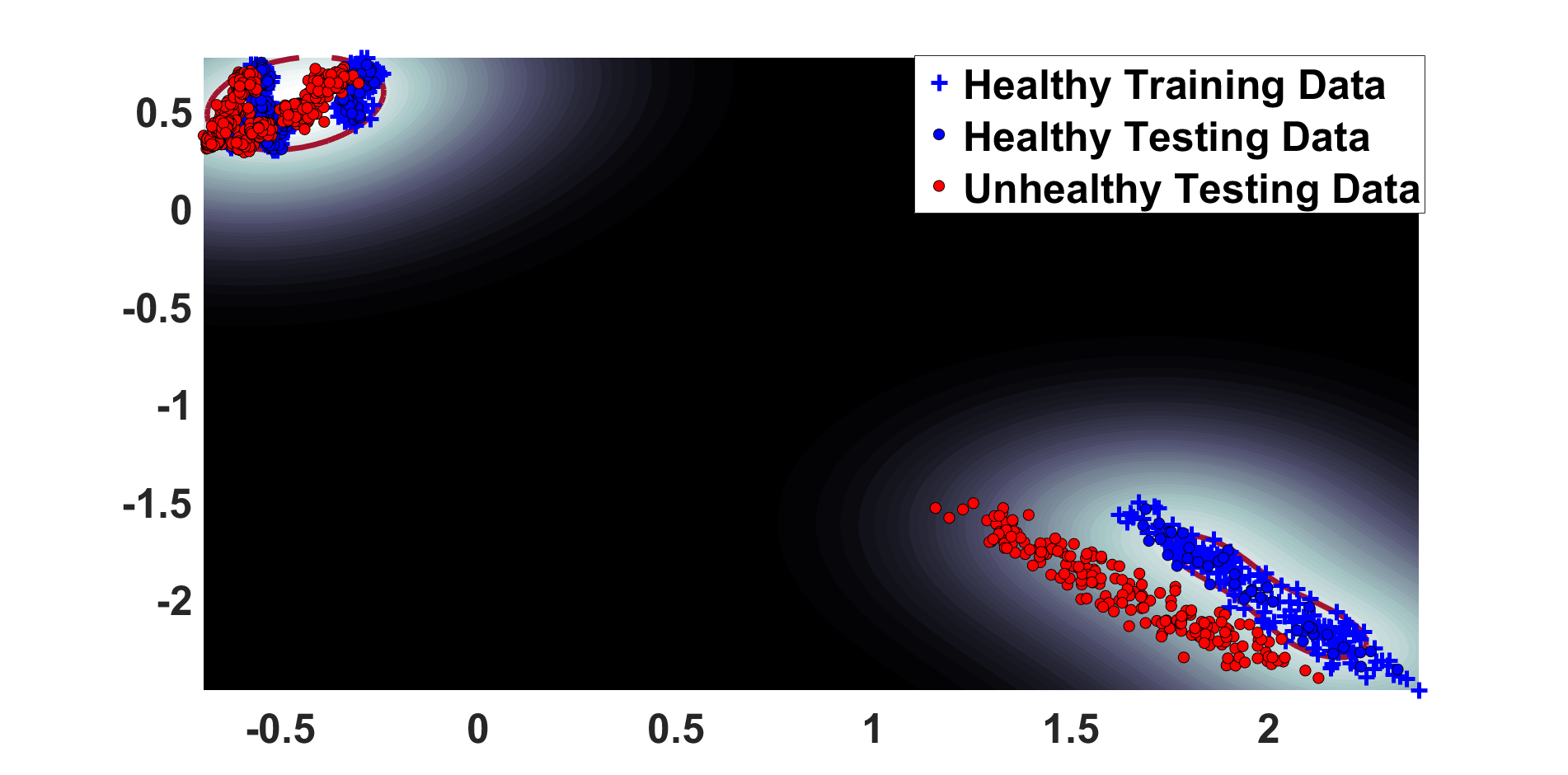}}
\subfloat[A zoomed in version of Figure \ref{fig:LHS_large_a} showing that clusters can be recovered if smaller input angles are neglected. \label{fig:LHS_large_b}]{\includegraphics[width=0.45\textwidth]{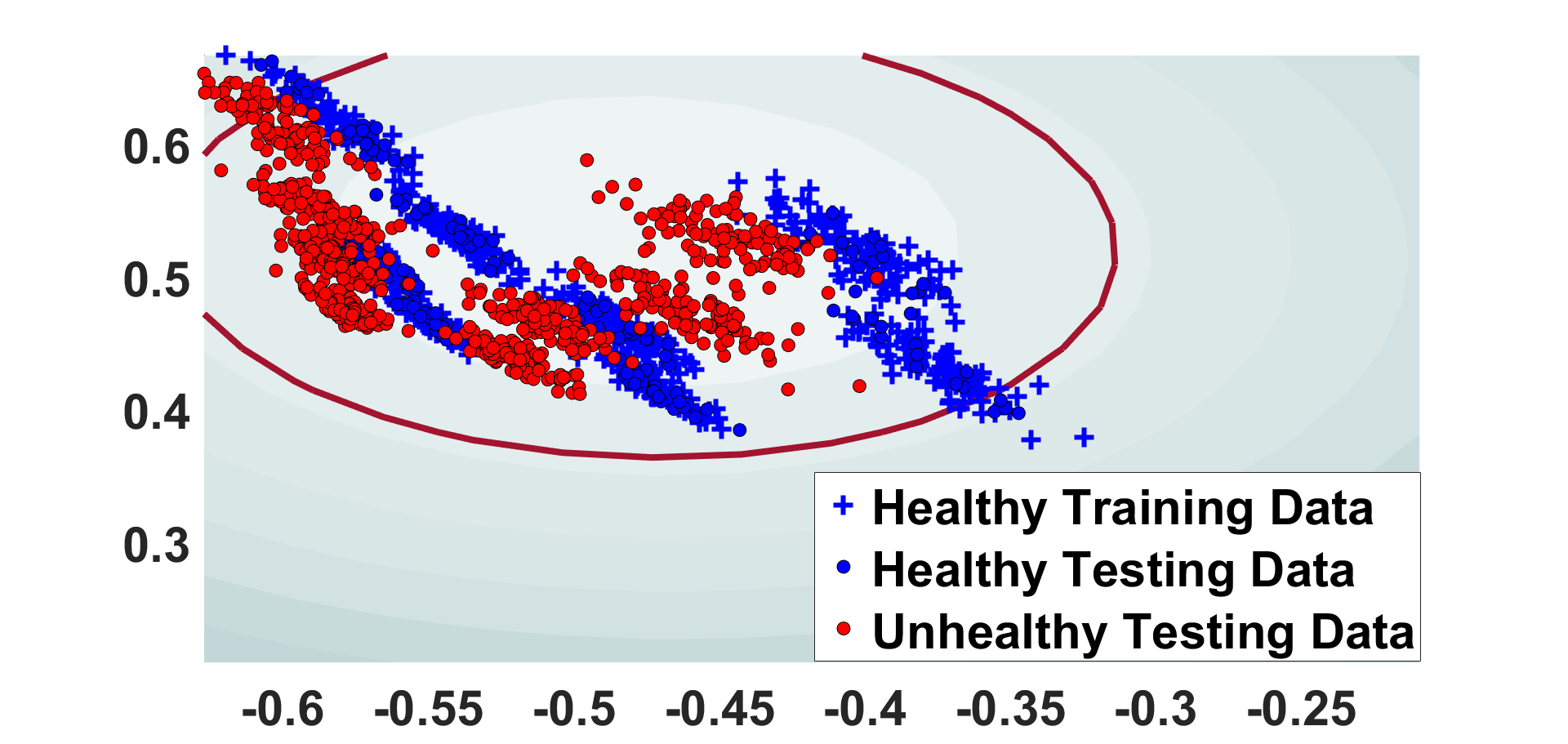}}
\caption{$\bm{C}$-space as a result of inputs being from an LHS, where each commanded input angle, $\beta_{C_1}$,$\beta_{C_2}\in \{[-8^{\circ},-5^{\circ}]\cup [5^{\circ},8^{\circ}] \}$.}
\label{fig:LHS_large}
\end{figure}

However, although the intrinsic clusters between the individual sensors have been lost (apart from sensor three against all the other four sensors), it appears that we are able to recover more structure if only the large angles are considered as input. That is we instead consider LHS samples, $\beta_{C_1}$,$\beta_{C_2}\in \{[-8^{\circ},-5^{\circ}]\cup [5^{\circ},8^{\circ}] \}$. This corresponds to the inputs shown previously in Figure \ref{fig:input_defl2}. Immediately we can see from Figure \ref{fig:LHS_large}, that there now once again exists implicitly generated clusters, which aids much more in interpretability, and also model building. It appears that CP decomposition tends to mix the behaviour of clusters together a lot more if only small magnitude input angles are considered. However large physical inputs allows for better separation in $\bm{C}$-space between the individual cluster correlated with each sensor, and between sensor three against the other sensors. However, although we have gained the \textit{qualitative} advantage of regrouping our data into clusters, it is still difficult for a one class SVM approach to work globally on this structure. This can be seen in Figure \ref{fig:LHS_large_b}, where SVM still cannot globally work on a per cluster basis, and from the previous analysis on artificial negative data generation, the closeness of the clusters will make it difficult for density-based approaches to prevent an overlapping between the clusters. This can be seen in Table \ref{tab:tab2_LHS}, where only a marginal increase in the $F_1$ scores are achieved. The purpose of the next section is to suggest that it is better to construct and individual one class SVM structure for each cluster, rather than a global model in order to improve predictive capability. 

\begin{table}[!h]
\centering
\begin{tabular}{ll|ll}
\multicolumn{2}{c|}{\textbf{All Angles}} & \multicolumn{2}{c}{\textbf{\begin{tabular}[c]{@{}c@{}}Large Angles\\ Only\end{tabular}}} \\ \hline
\multicolumn{1}{c}{946} & \multicolumn{1}{c|}{2054} & \multicolumn{1}{c}{320} & \multicolumn{1}{c}{680} \\
\multicolumn{1}{c}{47} & \multicolumn{1}{c|}{554} & \multicolumn{1}{c}{4} & \multicolumn{1}{c}{197} \\ \hline
$\nu$: & 0.07 & $\nu$: & 0.02 \\
g: & 0.8 & g: & 8 \\ \hline
$F_1$: & 0.34 & $F_1$: & 0.37
\end{tabular}
\caption{Confusion matrices, hyper-parameters, and scores for the one class SVM model in when considering LHS input data, with differing ranges of magnitudes. Damage data is considered as \textit{positive}, and undamaged data as \textit{negative} for the confusion matrix. \label{tab:tab2_LHS}}
\end{table}

\subsection{Considering Implicitly Generated Clusters}
  
 As the previous section has shown by considering the input space to consist only of large magnitude inputs, we are able to implicitly generate the clusters again. The reason for this is that low angles seem to introduce a lot of mixing effects in the CP decompositions, and physically speaking, larger angle inputs will result in larger accelerometer readings, allowing the signal pre-processing steps to be more effective in differentiating different events and sensors. However although different clusters were recovered, the one class SVM model was unable to construct and effective boundary around all the individual clusters, there by resulting in low metrics, such as the $F_1$-scores (and naturally all other binary classification metrics).
  
\begin{figure}[!h]
\centering
\subfloat[The cluster boundary strongly influenced by sensor three.]{\includegraphics[width=0.50\textwidth]{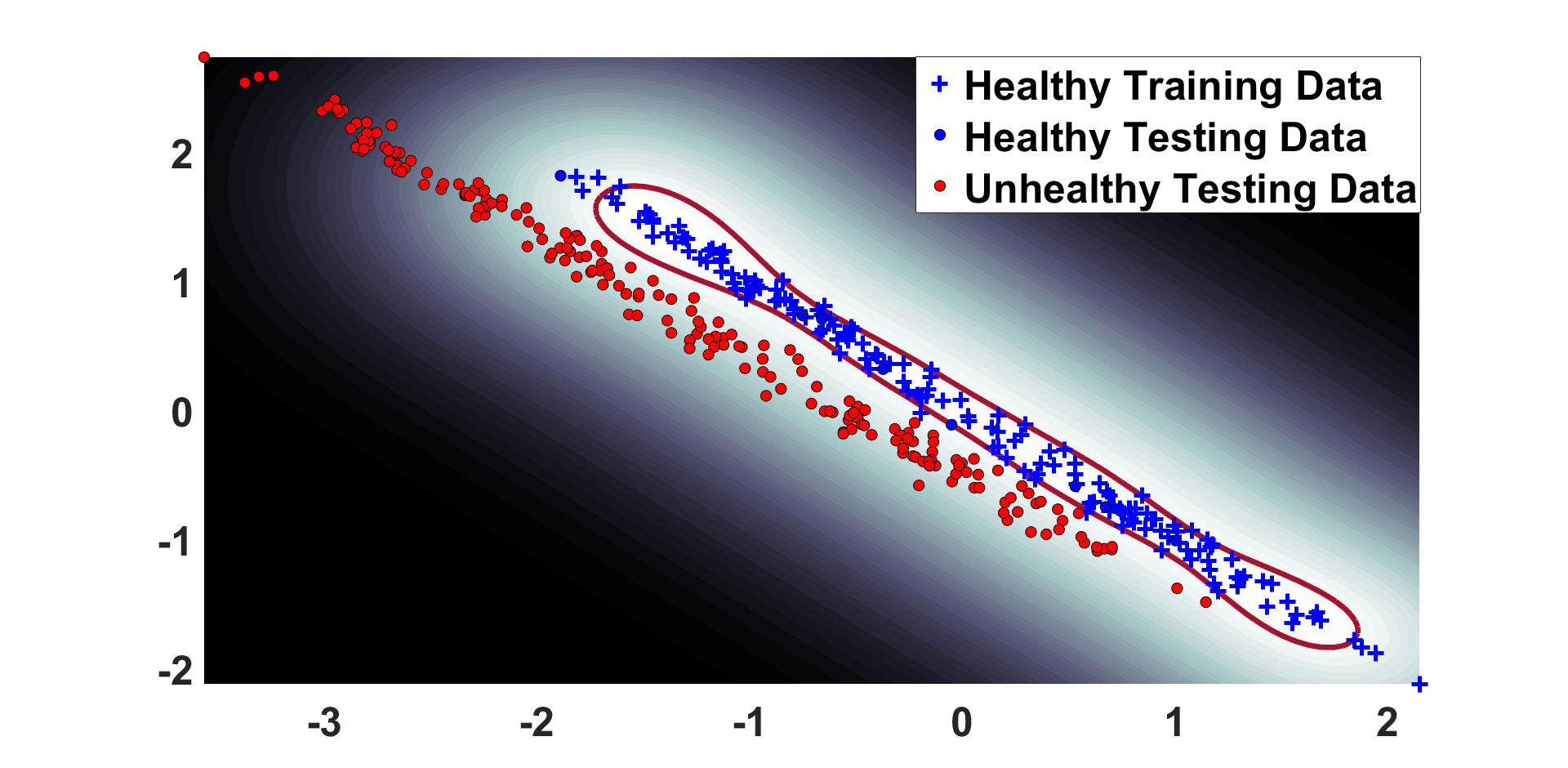}\label{fig:ind_clust_a}}
\subfloat[The cluster boundary strongly influenced by sensor five.]{\includegraphics[width=0.50\textwidth]{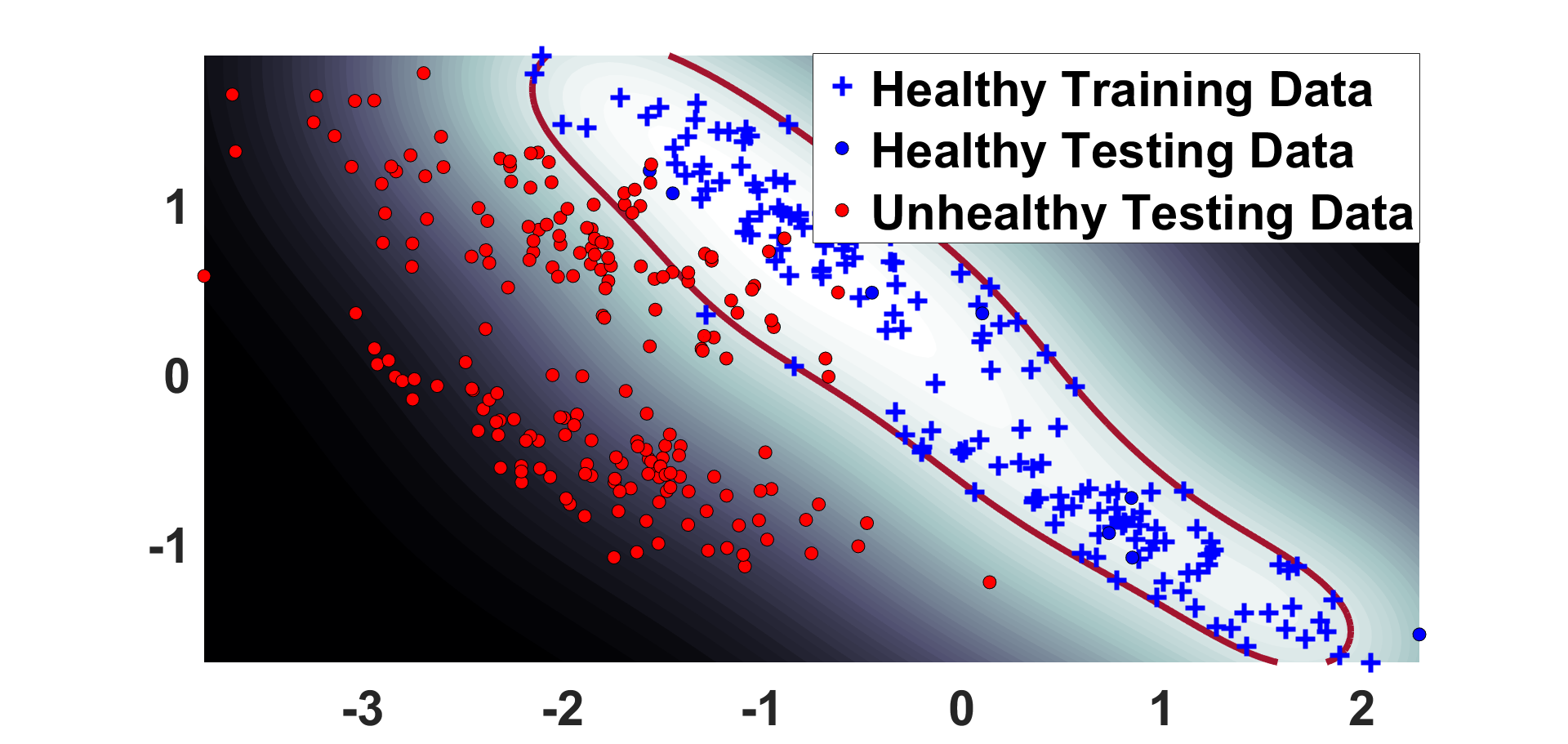}\label{fig:ind_clust_b}}
\caption{Differences in forming a one class SVM boundary around individual clusters, assuming large angle inputs generated via LHS.}
\label{fig:ind_clust}
\end{figure}

As a result, instead of trying to construct one large, \textit{global} one class SVM classifier, it was opted that instead multiple smaller classifiers be constructed around each individual cluster. In this way, the classification of new data points could be handled in any of number ways, for example by majority voting, or by weighted average between SVM boundaries, similar to the use of weighting coefficients for GMMs. Two example SVM boundaries are shown in Figure \ref{fig:ind_clust}, where Figure \ref{fig:ind_clust_a} refers to the cluster boundary strongly influenced by sensor three, and Figure \ref{fig:ind_clust_b} refers to the cluster boundary strongly influenced by sensor five. In both cases the boundary appears to be qualitatively better than that shown previously in Figure \ref{fig:LHS_large}, but as Table \ref{tab:tab3_ind} makes clear due to the low data count, the false negatives and positives which occur dramatically reduce the $F_1$-score of sensor three, when compared against sensor five. Moreover using more data will not help \textit{fix} this discrepancy, due to the physics behind sensor five's location. It lies half way long the chord and span, and so receives the smallest overall influence for wing twisting and bending. Therefore it can struggle to differentiate damage signals from healthy signals if the inputs are on the lower-end of the large magnitudes, or if the deflections of control surfaces result in an approximation modal phase cancellation at this centre point on the wing. Sensor five on the other experiences a large, direct influence from the commanded deflection angles resulting in its more robust boundary, and overall better score. However, as Table \ref{tab:tab3_ind} makes clear, it is possible to project sensor three into a higher dimensional space to increase the degree of separation between points, thereby giving better scores, as suggested in the previous section, \textit{Comparing Dimensionality}, it is preferable to try and avoid it if possible, due to the ability to introduce artificial noise if we project into too high of a dimension and are not careful (the reader is urged to read Bro \& Kiers to see if this may happen in their context, since notions of rank are not properly understand in tensor mathematics \cite{bro2003new}), and because for future SHM engineers models become much simpler to interpret if we can keep systems in a 2D space if possible, which according to this study is possible for ASE models.

\begin{table}[!h]
\centering
\begin{tabular}{ll|ll|ll}
\multicolumn{2}{c|}{\textbf{\begin{tabular}[c]{@{}c@{}}Cluster 5 Only \\ R = 2\end{tabular}}} & \multicolumn{2}{c|}{\textbf{\begin{tabular}[c]{@{}c@{}}Cluster 5 Only \\ R = 3\end{tabular}}} & \multicolumn{2}{c}{\textbf{\begin{tabular}[c]{@{}c@{}}Cluster 3 Only\\ R = 2\end{tabular}}} \\ \hline
\multicolumn{1}{c}{177} & \multicolumn{1}{c|}{23} & \multicolumn{1}{c}{200} & \multicolumn{1}{c|}{0} & 200 & 0 \\
\multicolumn{1}{c}{1} & \multicolumn{1}{c|}{7} & \multicolumn{1}{c}{1} & \multicolumn{1}{c|}{7} & 1 & 7 \\ \hline
$\nu$: & 0.05 & $\nu$: & 0.05 & $\nu$: & 0.05 \\
g: & 0.8 & g: & 0.8 & g: & 0.8 \\ \hline
$F_1$: & 0.37 & $F_1$: & 0.93 & $F_1$: & 0.93
\end{tabular}
\caption{Confusion matrices, hyper-parameters, and scores for the one class SVM model in when considering large angle LHS input data, with individual clustering. Damage data is considered as \textit{positive}, and undamaged data as \textit{negative} for the confusion matrix. Here \textit{Cluster} 3 and 5, correspond respectively to those clusters most strongly influenced by sensors 3 and 5 respectively. \label{tab:tab3_ind}}
\end{table}

\pagebreak
\section{Conclusion}

The field of SHM is extremely pertinent across all engineering disciplines.  SHM has been used successfully in the civil engineering field, and is currently in the phase of being slowly adopted by the aerospace field. Many complications do exist for the aerospace field, owing to the incredibly stringent factors of safety for aircraft, and due to a large difference in operating conditions for aircraft as opposed to civil structures, meaning that prior assumptions and algorithms developed in the civil SHM field need to be re-worked or re-modeled to work in aerospace. This was demonstrated by the construction of a Lagrangian $N$-DoF ASE model, which was derived from first principles in this paper, and subsequently used to study the ability for tensor, and one class SVM approaches to be used for SHM. It was found that certain assumptions in the civil SHM field do not readily apply to aerospace, namely due to the presence of active control surfaces, and the possibilities for larger bending and torsion modes to occur which result in multiple clusters for healthy and un-healthy data. However by projecting the clusters either into high-dimensional space, or by demanding large angle inputs and clustering these implicitly generated structure individually, it was found that standards in civil engineering were still able to be adopted to the aerospace engineering field. This demonstrates  an exciting future for this field, one that proves to be very fruitful in terms of research, and subsequently for industry.

\pagebreak
\bibliography{references_sci_tech_2019_extended_abs}

\appendix

\section{Full $N$ dof Aeroservoelastic Model}


\vspace{5cm}
\hspace{1cm}
\begin{sideways}
    \begin{minipage}{1\linewidth}
    \centering
      \includegraphics[width=1.3\linewidth,keepaspectratio]{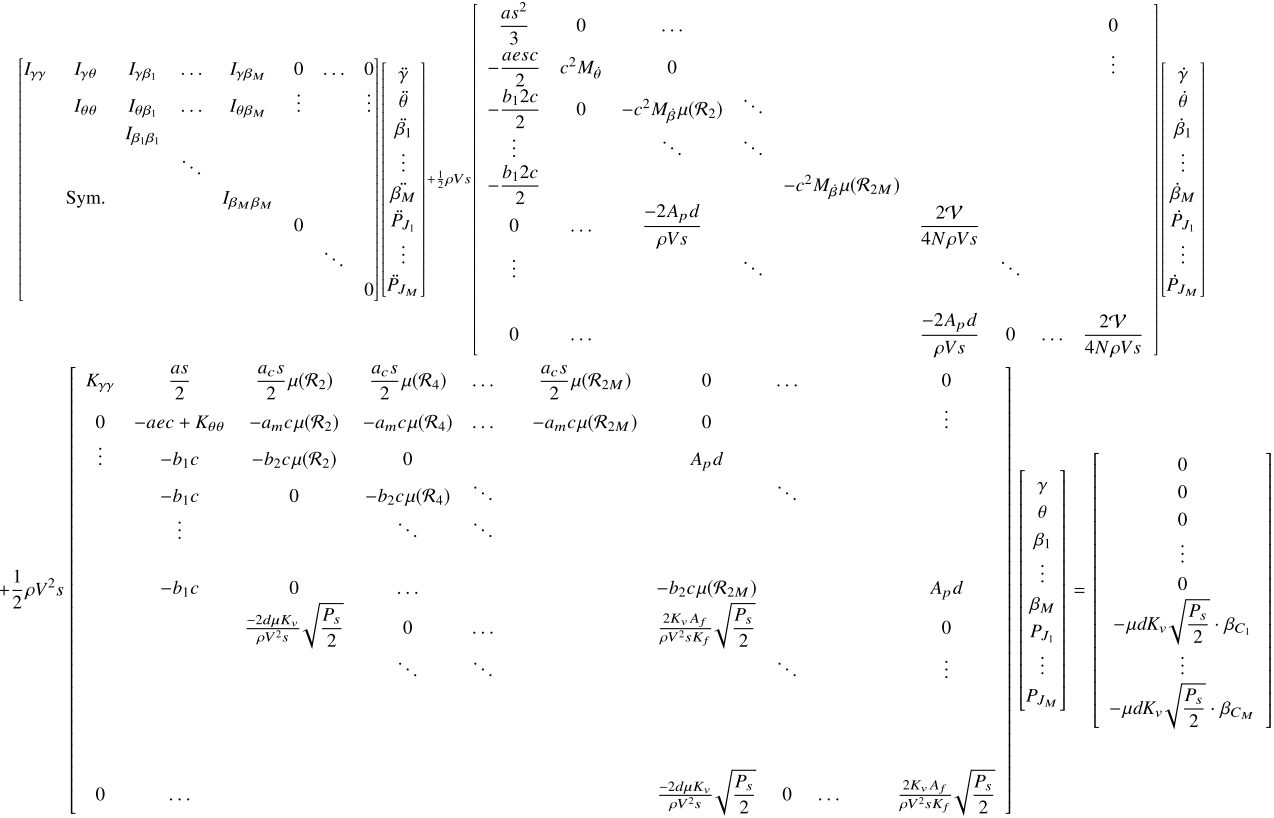}
           \begin{equation}
            \label{eqn:fkn_huuuuge}
            \end{equation}
    \end{minipage}
\end{sideways}

\end{document}